\documentclass{article}

\usepackage{PRIMEarxiv}

\usepackage{amsmath}
\usepackage{amssymb}
\usepackage{mathtools}
\usepackage{amsthm}
\usepackage{pifont}
\usepackage{tcolorbox}

\usepackage[utf8]{inputenc} 
\usepackage[T1]{fontenc}    
\usepackage{hyperref}       
\usepackage{url}            
\usepackage{booktabs}       
\usepackage{amsfonts}       
\usepackage{nicefrac}       
\usepackage{microtype}      
\usepackage{lipsum}
\usepackage{fancyhdr}       
\usepackage{graphicx}       
\usepackage{caption}
\usepackage{placeins}
\graphicspath{{media/}}     
\usepackage{enumitem}
\newcommand\mypara[1]{\vspace{1pt}\noindent\textbf{#1.}}

\title{3DXTalker: Unifying Identity, Lip Sync, Emotion, and Spatial Dynamics in Expressive 3D Talking Avatars}

\author{
  Zhongju Wang\thanks{Equal Contribution} \\
  University of New South Wales \\
  \texttt{zywang9691@gmail.com} \\
   \And
   Zhenhong Sun$^*$ \\
   Australian National University \\
  \texttt{zhenhongsun1992@outlook.com} \\
  \And
   Beier Wang \\
   University of New South Wales \\
  \texttt{beier.wang@unsw.edu.au} \\
  \And
   Yifu Wang \\
   Vertex Lab \\
  \texttt{usasuper@126.com} \\
  \And
   Daoyi Dong \\
   University of Technology Sydney \\
  \texttt{daoyidong@gmail.com} \\
  \And
   Huadong Mo\thanks{Corresponding Author} \\
   University of New South Wales \\
  \texttt{huadong.mo@unsw.edu.au} \\
  \And
   Hongdong Li \\
   Australian National University \\
  \texttt{hongdong.li@anu.edu.au} \\
}

\begin{document}
\maketitle

\vspace{-20pt}
\begin{center}
{\large \textbf{HomePage:} \href{https://engineeringai-lab.github.io/3DXTalker/}{ https://engineeringai-lab.github.io/3DXTalker/}}
\end{center}
\vspace{20pt}

\begin{abstract}
    Audio-driven 3D talking avatar generation is increasingly important in a wide range of multimedia applications. In these scenarios, avatars are expected to preserve identity, synchronize lip motion with speech, convey rich emotions, and exhibit flexible head pose, which together define the overall goal of expressivity. However, achieving this goal remains challenging due to insufficient diversity of  identities, expressions, and pose patterns in existing datasets, the absence of an integrated system, and limited semantic controllability.
    In this paper, we propose 3DXTalker,\footnote{We will open-source the complete data integration pipeline, processed dataset, and train-inference codebase for future research on expressive 3D talking avatars.} an integrated framework for expressive 3D talking avatar generation. First, we develop a scalable 2D-to-3D data integration pipeline that bridges large-scale videos and structured 3D facial representations, improving facial diversity. Second, we introduce a flow-matching framework in a disentangled parameter space, combining image-derived identity with audio-rich representations enhanced by frame-wise amplitude and emotion cues, to generate expressive 3D facial motions. Finally, we adopt a decoupled plug-in paradigm for semantic control over emotion and pose, enabling extensibility without disrupting the learned facial dynamics. Extensive experiments demonstrate that 3DXTalker achieves strong overall performance in identity preservation, lip synchronization, emotional expression, and controllable pose dynamics in 3D talking avatar generation.
\end{abstract}






\begin{figure}[t]
  \includegraphics[width=\textwidth]{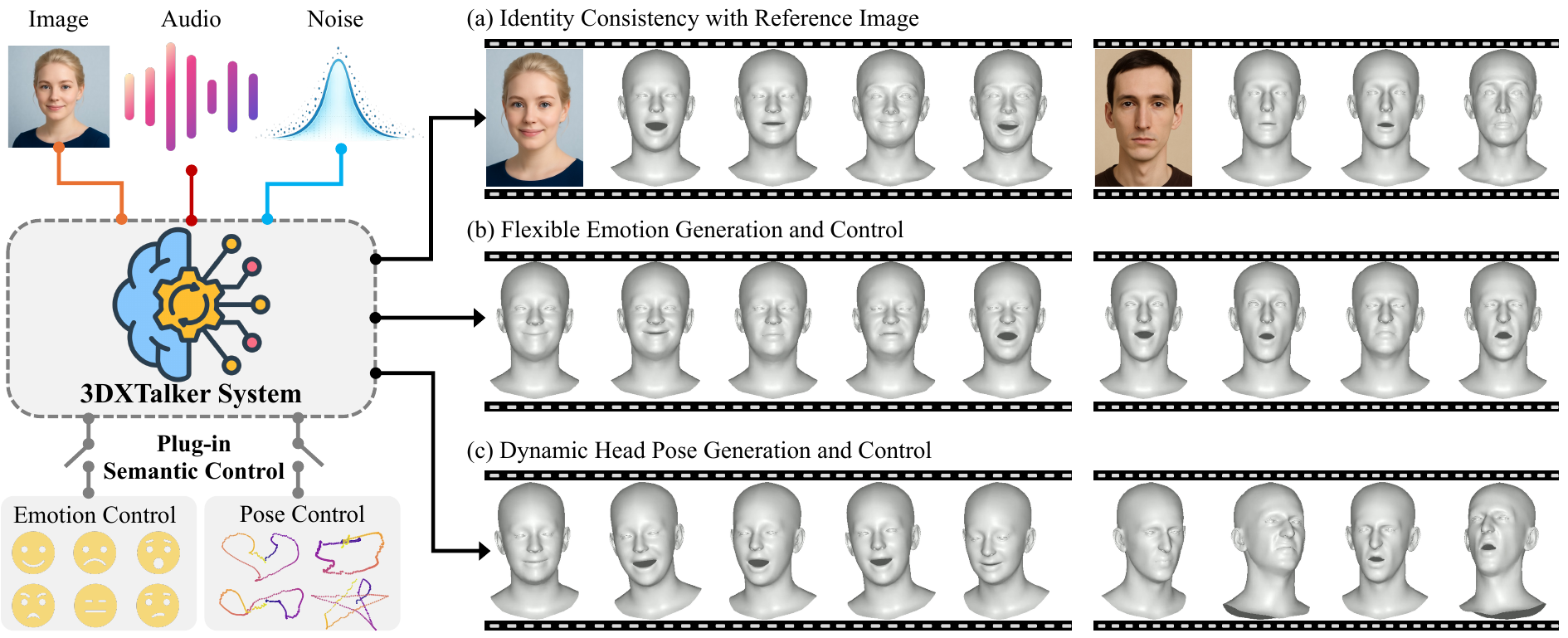}
  \caption{Overview of the proposed expressive 3DXTalker system with plug-in semantic control. Given a static reference image and a driving audio, 3DXTalker generates expressive 3D talking avatars with consistent identity, accurate lip synchronization, rich emotional expression, and controllable head pose dynamics. Demo videos are provided in the supplementary material.}
  \label{fig:teaser}
\end{figure}

\section{Introduction}
\label{sec:intro}
Audio-driven 3D talking avatars have been widely applied across various domains~\cite{volumetricheadavatars,3ddigitalhuman,singhead, instaglearningpersonalized3d} by mapping audio signals (speech or song) to realistic 3D facial movements, providing an effective way to animate avatars without complex 3D capture hardware. 
Early approaches~\cite{faceformer2022, VOCA2019, richard2021meshtalk} focused on producing lip movements, which underperformed against the emerging demand for personalized and expressive avatars. As applications evolve, this field has shifted toward a more comprehensive goal: achieving greater \textbf{Expressivity}, where avatars could preserve identity, synchronize lip motion with speech, convey emotional expressions, and exhibit pose dynamics, while supporting global semantic control, as illustrated in Figure~\ref{fig:teaser}.

Despite notable progress~\cite{danvevcek2023emotional, peng2023emotalk, kim2024deeptalk, sun2024diffposetalk}, achieving the goal of comprehensive expressivity in 3D talking avatars remains an ongoing challenge due to: 1) insufficient dataset coverage of diverse facial attributes; 2) the absence of an integrated framework for  joint modeling of identity consistency, emotional expression, and pose dynamics; and 3) limited semantic controllability for flexible global style modulation. 
Specifically, 3D audio–mesh datasets~\cite{VOCA2019, fanelli20103, wuu2022multiface, peng2023emotalk, wu2023mmface4dlargescalemultimodal4d} rely on costly real-world motion-capture devices, while current 2D-to-3D reconstructed datasets~\cite{sun2024diffposetalk, wu2024mmhead} lack diverse identities and facial detail, hindering generalization.
This limited coverage also impedes the creation of an integrated framework, while high-quality 3D avatars require a seamless fusion of expressivity from reference images and driving audio.
Moreover, real-world avatar systems require broader semantic controllability for style modulation (e.g., emotion and pose control), but the open-ended nature of such demands makes continuous data collection and retraining impractical.
These limitations hinder a model’s ability to capture the full spectrum of speech-driven expressivity for 3D avatars.

\begin{figure*}[t]
    \centering
    \includegraphics[width=1\linewidth]{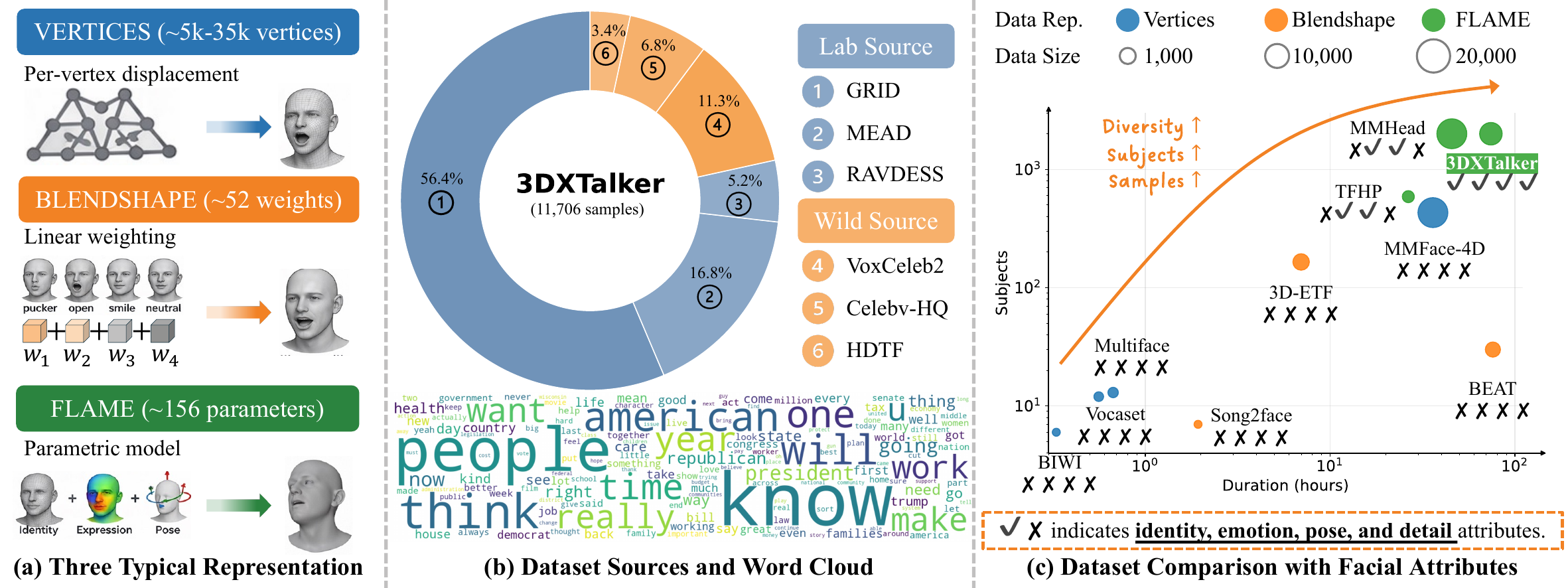}
    \caption{Overview of the proposed 3DXTalker dataset. (a) Comparison of three typical 3D talking-head representation paradigms. (b) Diverse sources of the 3DXTalker dataset from both lab and wild environments, together with a rich word-cloud text corpus. (c) Comparison of existing datasets in subject and sample scale, and attribute coverage, highlighting the suitability of the 3DXTalker dataset for expressive 3D talking avatar. Details of datasets are presented in \textbf{Appendix B.1}.}
    \label{fig:dataset}
\end{figure*}

Following 2D-to-3D lifting paradigms~\cite{sun2024diffposetalk, wu2024mmhead}, we seek to address insufficient dataset coverage by constructing a scalable bridge from large-scale 2D videos to structured 3D facial representations, analogous to the VAE in text-to-image generation. 
Built upon FLAME~\cite{FLAMESiggraphAsia2017} and DECA~\cite{DECASiggraph2021}, EMOCA~\cite{danvevcek2022emoca} recovers disentangled shape, expression, and pose parameters, along with fine facial details from video frames, making it suitable for our 3D facial modeling (Details in Part 1 of Sec.~\ref{sec:relatedWork}.).
Thus, we construct a 2D-to-3D data integration pipeline over three lab-controlled datasets and three in-the-wild datasets, with statistics in Figure~\ref{fig:dataset} (b) and (c). 
After unified filtering on duration, language, synchronization, and resolution, we build structured identity-aware FLAME representations via EMOCA and separate identity attributes from facial motions.
This yields a dataset with rich identity coverage, lip and emotional variations, flexible pose dynamics, without requiring expensive 4D scanning.

Utilizing the above dataset, we further introduce \textbf{3DXTalker}, an integrated flow-matching framework that operates in the disentangled parameter space to generate expressive 3D talking-head motion sequences conditioned on a reference image and driving audio. 
Beyond conventional audio embeddings for lip sync, we incorporate frame-wise amplitude features for coherent mouth aperture and frame-wise emotion features for nuanced expression modulation. This forms audio-rich representations that more faithfully reflect the dynamics of speech. The image-derived identity latent and audio-derived motion cues are jointly modeled via a multi-branch transformer, which produces identity-consistent, emotionally aligned talking-head motion. 
To further enhance semantic controllability, we leverage the disentangled flow-matching framework to adopt a decoupled plug-in paradigm, which is instantiated with two representative examples, emotion control and head-pose control at inference time, to demonstrate its extensibility while preserving learned facial dynamics. 
With these advances, 3DXTalker provides a unified paradigm for expressive 3D talking avatars, simultaneously addressing identity consistency, lip-sync accuracy, rich emotion, and flexible pose dynamics.

The main contributions of this paper are summarized as follows:
\begin{itemize}[leftmargin=*, noitemsep, nolistsep]
    \item[$\bullet$] We construct an identity-aware 3D talking dataset that bridges large-scale videos and structured 3D representations, improving attribute diversity to support more diverse avatars.
    \item[$\bullet$] We introduce a flow-matching framework with audio-rich representations that incorporate frame-wise amplitude and emotion features, improving holistic expression fidelity.
    \item[$\bullet$] We propose a decoupled plug-in paradigm for inference-time semantic control, which avoids retraining for open-ended demands and provides extensible controllability for 3D talking avatars.
\end{itemize}

\section{Related Work}
\label{sec:relatedWork}

\mypara{3D Facial Modeling and Datasets}
The goal of audio-driven 3D avatar synthesis is to generate realistic and expressive 3D facial motion from speech. While early methods mainly relied on expensive motion-capture setups~\cite{VOCA2019, fanelli20103, wuu2022multiface, peng2023emotalk, wu2023mmface4dlargescalemultimodal4d}, recent advances in monocular 2D-to-3D face reconstruction have enabled direct lifting from images and videos to structured 3D facial feature representations.
Existing methods mainly differ in their 3D facial representations, which can be broadly grouped into parametric and non-parametric forms.
Parametric models, such as 3DMM~\cite{booth20163d}, BFM~\cite{bfm2018}, and FLAME~\cite{FLAMESiggraphAsia2017}, represent facial geometry with low-dimensional shape, expression, and pose parameters, making them widely used in 3D facial modeling~\cite{DECASiggraph2021, danvevcek2022emoca, filntisis2022visual, zielonka2022mica}.
In particular, FLAME-based methods~\cite{retsinas20243d, zielonka2022towards, filntisis2023spectre, hempel20226d} can recover disentangled shape, expression, and pose parameters, providing a structured representation for 3D facial animation.
Building on this, EMOCA~\cite{danvevcek2022emoca} further models fine facial details beyond standard FLAME parameters, enabling more faithful identity preservation, which makes it especially suitable for expressive 3D talking avatar generation.
Non-parametric approaches~\cite{peng2025dualtalk, zhuang2025talkingeyes, wang2025ot, chu2025artalk} model 3D facial mesh animations by directly learning per-vertex displacement~\cite{VOCA2019, richard2021meshtalk} or blendshape linear weighting~\cite{peng2023emotalk, fan2024unitalker}. 
However, as shown in Figure~\ref{fig:dataset} (a), such representations often remain high-dimensional or semantically entangled, which makes fine-grained motion control more difficult despite their strong deformation flexibility.
More importantly, existing 2D-to-3D reconstructed datasets~\cite{sun2024diffposetalk, wu2024mmhead} still lack sufficient comprehensive coverage of both identity-related geometry and fine facial details for expressive 3D talking avatars. We therefore adopt EMOCA as our 2D-to-3D bridge, playing a role similar to a VAE in text-to-image generation by encoding input facial frames into a compact yet expressive latent representation that supports expressive avatar generation.

\mypara{Audio-driven Regression Models}
Early methods mainly formulate audio-driven 3D facial animation as a regression problem, where speech features are deterministically mapped to facial motion. These approaches typically employ features extracted from large-scale self-supervised speech models, such as Wav2vec 2.0~\cite{baevski2020wav2vec}, HuBERT~\cite{hsu2021hubert}, and WavLM~\cite{chen2022wavlm}, and regress either raw vertex displacements~\cite{faceformer2022, scantalk, peng2023selftalk, renjie2024audio} or parametric model latents~\cite{kim2024deeptalk, LearningToListen, laughtalk, fan2024unitalker, xing2023codetalker, peng2023emotalk, danvevcek2023emotional, shen2024deitalk}.
FaceFormer~\cite{faceformer2022} predicts vertex trajectories using a transformer decoder, while UniTalker~\cite{fan2024unitalker} and CodeTalker~\cite{xing2023codetalker} compress facial motion into low-dimensional latents for efficient learning. Although regression models can achieve good lip synchronization, their deterministic formulation is less effective at modeling the inherently one-to-many nature of speech-driven facial motion, often leading to limited expressivity and reduced facial dynamics that generative modeling can better support.

\mypara{Audio-driven Generative Models}
To improve the diversity and realism of audio-driven 3D facial animation, recent works increasingly adopt generative formulations that model the conditional distribution from speech audio to facial motion. In particular, diffusion-based approaches (e.g., FaceDiffuser~\cite{stan2023facediffuser}, DiffPoseTalk~\cite{sun2024diffposetalk}, DiffusionTalker~\cite{chen2023diffusiontalker}, FaceTalk~\cite{aneja2024facetalk}) generate motion by progressively denoising from Gaussian noise, yielding more diversity and realism. 
Among them, some works mainly focus on lip synchronization, while others emphasize emotional facial expressions or head-pose dynamics separately. For instance, EMOTE~\cite{danvevcek2023emotional} focuses on modeling emotional facial expressions, while DiffPoseTalk~\cite{sun2024diffposetalk} explores audio-driven head-pose dynamics as a separate component. 
Despite these improvements in expressivity, existing methods still lack an integrated framework that jointly addresses identity consistency, expression, pose, and detail, especially for unseen subjects. 
Therefore, we present an integrated flow-matching framework with audio-rich representations and decoupled plug-in semantic control for expressive 3D talking avatar generation.


\section{Method}
\label{sec:methodology}
In the task of audio-driven 3D talking avatar generation, the core objective is to synthesize a sequence of 3D avatar states $\{\mathbf{M}_i\}_{i=1}^N$ ($N$ total frames) that align with a given audio waveform $\mathbf{A}$, ensuring precise synchronization between facial movements and speech. 
Beyond lip synchronization, expressive avatar generation additionally requires identity preservation, emotion-aware facial expressions, and natural pose dynamics. However, this remains challenging due to limited training data with diverse facial attributes, the lack of an integrated framework for expressive facial dynamics, and restricted semantic controllability. To address these issues, we propose 3DXTalker, which improves expressive avatar generation through data integration and reformulation, the integrated flow-matching framework, and decoupled plug-in semantic control as follows.
\begin{figure*}[t]
    \centering
    \includegraphics[width=\linewidth]{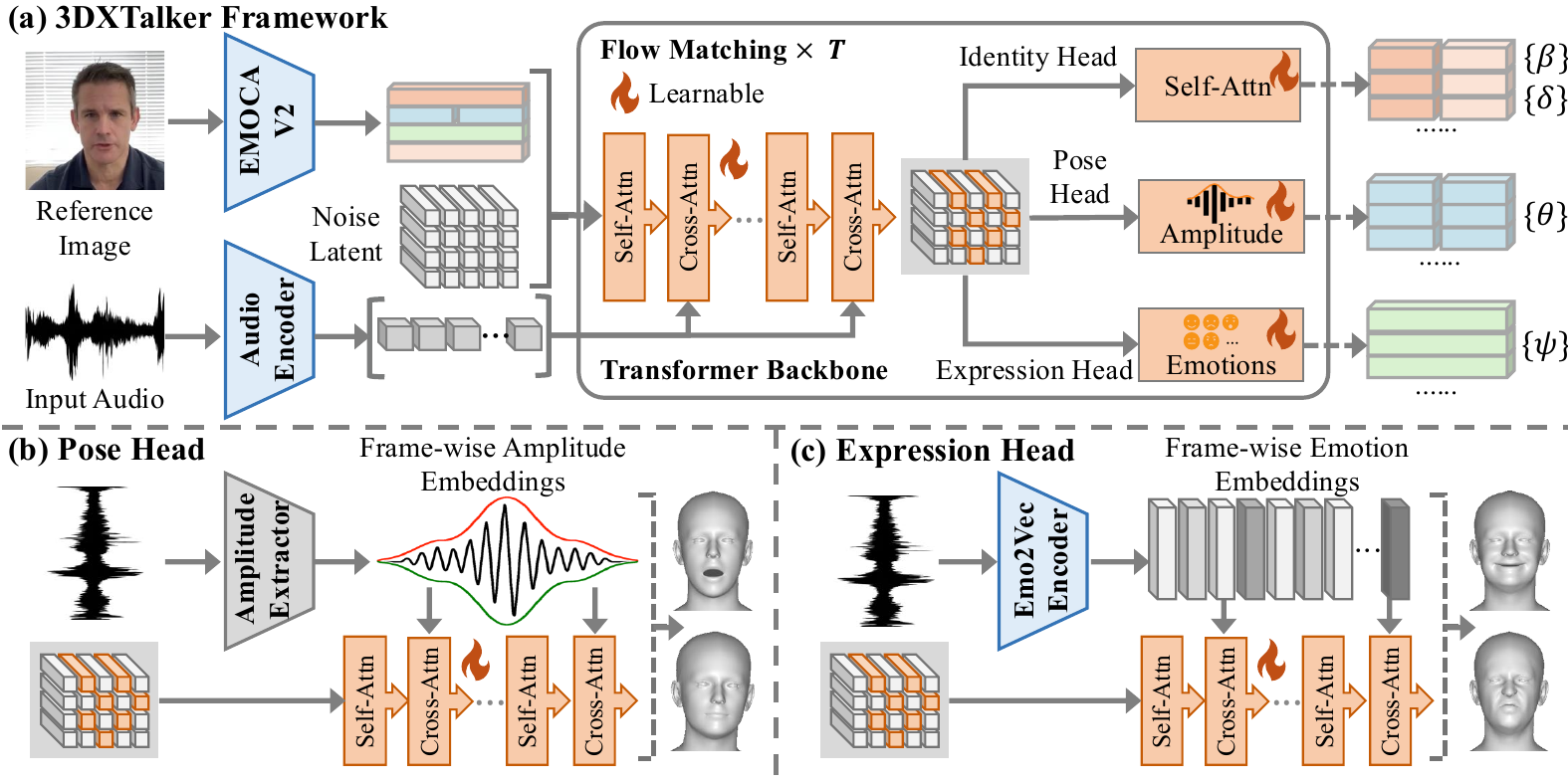}
    \caption{Overview of 3DXTalker framework. 
    (a) A multi-branch flow-matching transformer fuses identity and audio cues to model disentangled FLAME parameter space.
    (b) Frame-wise audio amplitude contributes to coherent mouth aperture and head dynamics.
    (c) Frame-wise emotion embeddings help modulate emotional expressions.
    }
    \label{fig:Overview_xtalker}
\end{figure*}

\subsection{Data Integration and Reformulation}
\label{subsec:id}
\mypara{Scalable Data Integration}
To overcome the limited attribute coverage of existing 3D talking-avatar datasets while avoiding expensive 3D capture, we built a scalable data integration pipeline that unifies diverse video sources into a structured 3D facial representation space.
As illustrated in Figure~\ref{fig:dataset} (b) and (c), we integrate three lab-controlled datasets (GRID~\cite{cooke2006audio}, RAVDESS~\cite{livingstone2018ryerson}, MEAD~\cite{kaisiyuan2020mead}) and three in-the-wild datasets (VoxCeleb2~\cite{chung18b_interspeech}, HDTF~\cite{zhang2021flow}, CelebV-HQ~\cite{zhu2022celebvhq}) into a shared 3D representation space. The lab-controlled datasets provide clean articulation and explicit emotional variation, while the in-the-wild datasets contribute broader identity diversity, speaking styles, and natural pose dynamics.

To improve reliability and consistency across heterogeneous sources, we apply a unified filtering pipeline that consists of: (1) duration thresholding to retain or stitch clips longer than 10 seconds, (2) signal-to-noise ratio filtering to suppress noisy or corrupted speech, (3) language filtering to maintain linguistic consistency, (4) audio--visual synchronization verification to remove misaligned segments, and (5) spatial resolution normalization to $512\times512$. Through this process, large-scale 2D videos are converted into a structured corpus suitable for expressive 3D talking-avatar modeling.
Further details are provided in \textbf{Appendix B.1}. 

\mypara{Identity-Aware Differential Reformulation}
After data integration, each video frame is lifted into the FLAME parameter space using the EMOCA encoder~\cite{danvevcek2022emoca}, yielding a structured representation, including shape $\boldsymbol{\beta} \in \mathbb{R}^{100}$, pose $\boldsymbol{\theta}\in \mathbb{R}^{6}$, expression $\boldsymbol{\psi} \in \mathbb{R}^{50}$ and an additional detail parameter $\boldsymbol{\delta} \in \mathbb{R}^{128}$. Notably, the pose $\boldsymbol{\theta}$ can be decomposed into \emph{head pose} and \emph{jaw pose}. These parameters can be decoded into a coarse mesh $\mathbf{M}_{coa} \in \mathbb{R}^{5023 \times 3}$ deformed from a FLAME template $\mathbf{\bar{V}}$ with $\boldsymbol{\beta}$, $\boldsymbol{\psi}$, and $\boldsymbol{\theta}$, which is refined by the detail decoder ${\mathcal{D}_{det}}$ to form a detailed mesh $\mathbf{M}_{det} \in \mathbb{R}^{59315 \times 3}$ through a facial displacement map, formally defined as:
\begin{equation}
    \begin{aligned}
    \mathbf{V}_{int} &= \mathbf{\bar{V}} + \text{B}_S(\boldsymbol{\beta}) + \text{B}_E(\boldsymbol{\psi}), \\
    \mathbf{M}_{coa} &= \text{LBS}(\mathbf{V}_{int}, \text{J}_P(\mathbf{\bar{V}}), \mathcal{W}, \boldsymbol{\theta}),\\
    \mathbf{M}_{det} &= \text{F}_{det}(\mathbf{M}_{coa}, (\mathbf{\bar{U}}+{\mathcal{D}_{det}}(\boldsymbol{\theta}, \boldsymbol{\psi}, \boldsymbol{\delta}))),
    \end{aligned}
    \label{eq:flame}
\end{equation}
where $\mathbf{V}_{int}$ is an intermediate mesh, $\text{B}_S$ and $\text{B}_E$ are shaping and expressing functions, $\text{LBS}$ denotes the linear blend skinning algorithm with joint regressors $\text{J}_P$ and skinning weights $\mathcal{W}$, $\mathbf{\bar{U}}$ denotes template UV map, and $\text{F}_{det}$ applies the predicted displacement map to the coarse mesh $\mathbf{M}_{coa}$. Details are provided in \textbf{Appendix A}.

Rather than modeling these frame-wise parameters directly, we further reformulate them into an identity-anchored differential representation to separate stable identity attributes from temporally varying facial motion. Concretely, we take the first frame as a reference and express subsequent frames in relative form:
{\small
\begin{equation}
\mathbf{X}_\Delta = \left\{ 
    \left(
        \boldsymbol{\beta}_i - \boldsymbol{\beta}_0,\;
        \boldsymbol{\delta}_i - \boldsymbol{\delta}_0,\;
        \boldsymbol{\psi}_i - \boldsymbol{\psi}_0,\;
        \boldsymbol{\theta}_i - \boldsymbol{\theta}_0
    \right)\right\}_{i=1}^{N}. 
\label{eq:abs}
\end{equation}
}
This representation enables identity stability while lip movements, expressions, and head dynamics evolve over time. The resulting parametric sequence thus forms a consistent, controllable, and compact 3D facial motion trajectory, which serves as the basis for the subsequent generative modeling in our framework.

\subsection{Integrated Flow-matching Framework}
\label{subsec:framework}

The data integration and reformulation pipeline obtains disentangled identity and motion representations from 2D videos using EMOCA. This allows a single reference image to serve as a reliable identity anchor during generation, in contrast to prior 3D audio-driven methods that relied on a fixed identity template. Additionally, the audio side remains a limiting factor. Speech signals inherently contain multiple layers of information: (1) linguistic content (word, phoneme), (2) articulatory dynamics reflected in amplitude and rhythm that drive jaw motion and mouth aperture, and (3) emotional prosody conveyed through intonation, energy contours, and vocal timbre. Common audio embeddings focus on linguistic content~\cite{baevski2020wav2vec, hsu2021hubert, chen2022wavlm}, but overlook prosodic cues, leading to correct word synchronization but weak mouth aperture and flat emotion expressions.

\mypara{Integrated audio-rich Backbone}
To address the aforementioned gap, we introduce an integrated flow-matching framework that operates in the disentangled parameter space to generate expressive 3D talking-head sequences conditioned on a reference image $\mathbf{I}_0$ and driving audio $\mathbf{A}$. Beyond conventional audio embeddings, we incorporate frame-wise amplitude features for coherent mouth aperture and frame-wise emotion features for nuanced expression modulation, forming \textbf{audio-rich representations} that more precisely reflect the dynamics of speech. The identity latent derived from the reference image and the audio cues is then jointly modeled through a multi-branch flow-matching transformer, enabling identity consistency, lip synchronization, and emotional alignment in talking-head motion, as depicted in Figure~\ref{fig:Overview_xtalker}. We next provide the detailed process of this framework.

Given an input image $\mathbf{I}_0$, we first extract the reference parametric representation $\mathbf{X}_{ref}=(\boldsymbol{\beta}_0,\boldsymbol{\theta}_0,\boldsymbol{\psi}_0, \boldsymbol{\delta}_0) \in \mathbb{R}^{1\times284}$, which is then combined with $t$ step-dependent noise $\boldsymbol{\varepsilon}_{t} \in \mathbb{R}^{N \times 284}$ (initial random noise $\boldsymbol{\varepsilon}_{0}$) to form the model’s input state $\mathbf{\tilde{X}}_t$ after the MLP layers. The input audio $\mathbf{A}$ is processed into linguistic embeddings $\mathbf{A}_{feat}\in \mathbb{R}^{N \times d}$ ($d$ represents the dimension) using WavLM~\cite{chen2022wavlm}. In the Transformer backbone $\mathcal{D}_{flow}$, $\mathbf{\tilde{X}}_t$ serves as queries, while the audio embeddings $\mathbf{A}_{feat}$ act as keys and values. Through self-attention and cross-attention, the model produces the intermediate latent representation $\mathbf{H}_{t}$. The overall process can be summarized:
\begin{equation}
    \begin{aligned}
    \mathbf{\tilde{X}}_t &= \text{MLP}(\boldsymbol{\varepsilon}_{t}) + \text{MLP}(\mathbf{X}_{ref}), \\
    \mathbf{H}_{t} &= \mathcal{D}_{flow}(\mathbf{\tilde{X}_t}, \mathbf{A}_{feat}, t).
    \end{aligned}
    \label{eq:dit}
\end{equation}
$\mathbf{H}_{t}$ serves as the fused latent representation that integrates the reference identity and the linguistic content from audio and is subsequently passed through three parallel branches to disentangle and predict different FLAME parameters.

This branch is responsible for predicting the shape $\{\boldsymbol{\beta}\}$ and detail $\{\boldsymbol{\delta}\}$ parameters that define the identity of the generated talking head. With the global latent representation $\mathbf{H}_{t}$ extracted, the identity head employs lightweight self-attention layers $\mathcal{D}_{id}$ to further form identity-related features and subsequently passes them through an MLP layer to produce the velocity field of shape and detail ($\hat{\boldsymbol{v}}^{\beta}_t \in \mathbb{R}^{N \times 100}$  and $\hat{\boldsymbol{v}}^{\delta}\in \mathbb{R}^{N \times 128}$), defined as:
\begin{equation}
(\hat{\boldsymbol{v}}^{\beta}_t;\hat{\boldsymbol{v}}^{\delta}_t)= \text{MLP}(\mathcal{D}_{id}(\mathbf{H}_{t})).
    \label{id_head}
\end{equation}

To achieve precise and responsive control of jaw motion and head rotation, we extract frame-wise amplitude features $\mathbf{A}_{amp}$ from the driving audio. Specifically, we first compute the amplitude envelope of the waveform via the Hilbert transform and then apply frame-level window averaging to obtain a temporally aligned amplitude sequence that reflects speech intensity and rhythmic variations. These amplitude cues are injected into the pose branch via a cross-attention module $\mathcal{D}_{pose}$, where $\mathbf{H}_{t}$ provides motion context and $\mathbf{A}_{amp}$ provides audio-driven modulation. The resulting features are decoded by an MLP to produce the velocity fields for jaw pose $\hat{\boldsymbol{v}}^{\theta^{j}}_t \in \mathbb{R}^{N \times 3}$ and global head rotation $\hat{\boldsymbol{v}}^{\theta^{g}}_t \in \mathbb{R}^{N \times 3}$:
\begin{equation}
(\hat{\boldsymbol{v}}^{\theta^{j}}_t,\; \hat{\boldsymbol{v}}^{\theta^{g}}_t)
= \text{MLP}\big(\mathcal{D}_{pose}(\mathbf{H}_{t}, \mathbf{A}_{amp})\big).
\label{pose_head}
\end{equation}

To enable emotionally coherent facial expressions, we extract frame-wise emotion embeddings $\mathbf{A}_{emo}$ from the driving audio using emotion2vec~\cite{ma2023emotion2vec}. These embeddings capture subtle affective cues (e.g., happiness, sadness, anger) embedded in speech and are temporally aligned with the latent representation. The expression head injects $\mathbf{A}_{emo}$ into the motion representation via a cross-attention module $\mathcal{D}_{exp}$, followed by an MLP that predicts the expression velocity field $\hat{\boldsymbol{v}}^{\psi}_t \in \mathbb{R}^{N \times 50}$, represented as:
\begin{equation}
\hat{\boldsymbol{v}}^{\psi}_t
= \text{MLP}\big(\mathcal{D}_{exp}(\mathbf{H}_{t}, \mathbf{A}_{emo})\big).
\label{exp_head}
\end{equation}

\mypara{Flow-matching Training and Inference}
To improve optimization stability in training, we initialize the latent trajectory using a temporally smoothed noise sequence $\boldsymbol{\varepsilon}_0 \in \mathbb{R}^{T \times D}$. A sparse set of anchor latents is first sampled and then linearly interpolated to form a smooth trajectory. The anchor locations are adaptively determined by extrema of the speech amplitude envelope rather than fixed temporal intervals, introducing a weak articulation-related prior that improves temporal coherence and stabilizes mouth-aperture dynamics.
Then, we model the continuous evolution of FLAME parameters as a straight flow between the initial latent $\boldsymbol{\varepsilon}_0$ and the target displacement $\mathbf{X}_\Delta = \mathbf{X} - \mathbf{X}_{ref}$. 
Given a randomly sampled $t \sim \mathcal{U}(0,1)$, the target velocity is defined as the linear interpolant, and the flow matching objective supervises the model-predicted velocity $\hat{\boldsymbol{v}}(t)$ as follows:
\begin{equation}
\mathcal{L}_{\text{flow}}
=
\mathbb{E}_{t \sim \mathcal{U}(0,1)}
\left[
\left\|
\hat{\boldsymbol{v}}(t)
-
\big(t\,\mathbf{X}_\Delta + (1-t)\,\boldsymbol{\varepsilon}_0\big)
\right\|_2^2
\right].
\label{eq:flow_loss}
\end{equation}
This formulation directly aligns the model with the continuous flow that transports the latent state toward the target parameter space, enabling stable optimization, temporally smooth synthesis, and coherent evolution across different facial attributes over time. 

In inference, the velocity fields from all heads are concatenated ($\boldsymbol{\hat{v}}_t=\text{Concat}(\,\hat{\boldsymbol{v}}^{\beta}_t,\; \hat{\boldsymbol{v}}^{\psi}_t,\; \hat{\boldsymbol{v}}^{\theta}_t,\; \hat{\boldsymbol{v}}^{\delta}_t\,)$) to form a unified displacement field $\boldsymbol{\varepsilon}{t}$, which is updated iteratively for $T{\text{inf}}$ steps from the initial latent state. The final FLAME parameters are then recovered by adding the predicted displacement to the reference representation:
\begin{equation}
\hat{\mathbf{X}} = {\mathbf{X}}_{ref} + \boldsymbol{\varepsilon}_{T_{\text{inf}}}, \quad
\boldsymbol{\varepsilon}_{t} = \boldsymbol{\varepsilon}_{t-1} + 1/T_{inf}\times \boldsymbol{\hat{v}}_t.
\label{eq:flow_infer}
\end{equation}

Overall, the proposed audio-rich flow-matching framework provides a unified approach to model identity and speech-driven facial dynamics in the disentangled FLAME parameter space.

\subsection{Decoupled Plug-in Semantic Control}
\label{subsec:optimization}

Built upon the proposed dataset and framework, our system already enables expressive and identity-consistent 3D talking avatar generation. However, in real-world scenarios, avatars often require additional global controllability to adapt their delivery styles to different contexts, such as varying emotional intensity or modulating head-pose motion patterns. 
Although specific control factors can be improved with more targeted data and annotations, broader global stylistic demands are inherently diverse and difficult to enumerate, making it impractical to absorb all such controllability into flow-matching training. 
Instead, benefiting from the disentangled flow-matching framework, we adopt a decoupled plug-in paradigm that introduces emotion and head-pose control at inference time while preserving learned facial dynamics, and is readily extensible to broader global semantic controls beyond these two cases.

Benefiting from the disentanglement of shape and expression parameters in FLAME, we construct a set of semantic expression templates $\{\bar{\boldsymbol{\psi}}^{e}\}_{e=1}^7$ from the MEAD dataset, covering seven emotional styles: Angry, Contempt, Disgust, Fear, Happy, Sad, and Surprise. Each template is associated with an intensity factor $\alpha \in \{1, 1.2, 1.4, 1.6, 1.8, 2.0\}$ to control the strength of its intrinsic emotional semantics. During inference, we modulate the global emotional semantics by interpolating between the reference expression $\boldsymbol{\psi}_{ref}$ and the scaled emotion template:
\begin{equation}
\boldsymbol{\hat{\psi}}_{ref}^{\,e}
=
(1-\lambda)\,\boldsymbol{\psi}_{ref}
+
\lambda\,\alpha_e\, \bar{\boldsymbol{\psi}}^{e}.
\end{equation}
This design provides seven emotion categories, each with six controllable intensity levels, while preserving audio-driven expression dynamics. More details are in \textbf{Appendix D.1}.

Complementary to emotion semantic control, we further introduce semantic control over head-pose dynamics. Given the driving audio or a text prompt describing the desired presentation style, a language model generates a stable semantic head-pose trajectory, such as gentle sways, rhythmic arcs, or gradual rotations. This trajectory is superimposed onto the model-predicted natural head motion rather than replacing it, which preserves realism while maintaining stable temporal transitions. In this way, the model supports semantically meaningful stylistic variation, such as calm or energetic delivery, while preserving motion coherence and natural temporal continuity. Details are in \textbf{Appendix E.1}.


\begin{figure*}[t]
    \centering
    \includegraphics[width=0.95\linewidth]{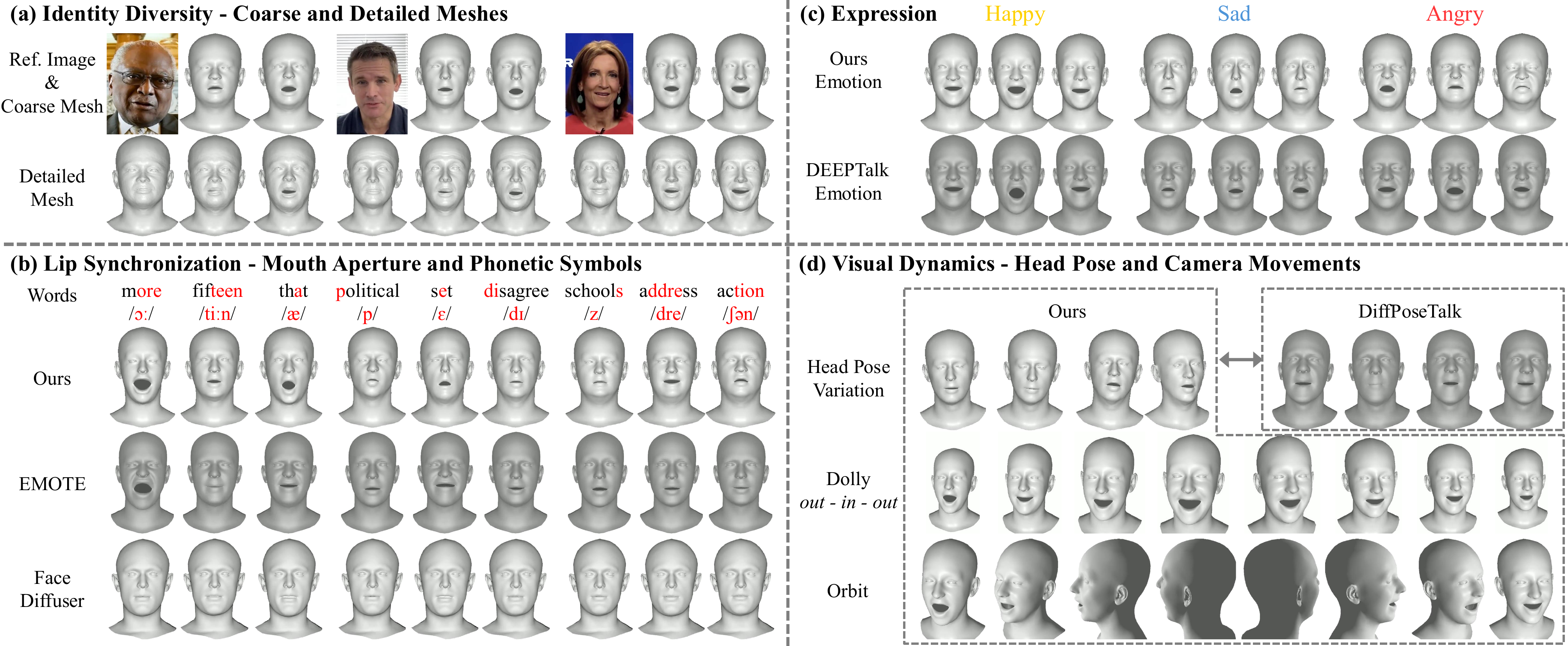}
    \caption{Qualitative comparisons over selected typical baselines. 
    (a) shows the consistency between generated meshes and the reference image.
    (b) shows better mouth aperture alignment.
    (c) shows finer emotional expressiveness.
    (d) shows predicted natural head pose and camera movements.
    Full baseline comparisons are provided in \textbf{Appendix C}, and more emotion comparisons are offered in \textbf{Appendix D.2}. We further present downstream applications of 3DXTalker in \textbf{Appendix G}.}
\label{fig:QualitativeComparison}
\end{figure*}

\section{Experiments}
\label{sec:exp}

\subsection{Implementation Details}
\label{subsec:details}
\mypara{Baselines}
We compare 3DXTalker with seven representative baselines spanning diverse paradigms for audio-driven 3D talking-avatar generation. The evaluated baselines are: (1) FaceFormer~\cite{faceformer2022}; (2) CodeTalker~\cite{xing2023codetalker}; (3) SelfTalk~\cite{peng2023selftalk}; (4) DiffPoseTalk~\cite{sun2024diffposetalk}; (5) EMOTE~\cite{danvevcek2023emotional}; (6) FaceDiffuser~\cite{stan2023facediffuser}; and (7) DEEPTalk~\cite{kim2024deeptalk}. 

\mypara{Datasets}
We train models on six diverse talking-head datasets, covering lab-scanned (GRID~\cite{cooke2006audio}, RAVDESS~\cite{livingstone2018ryerson}, MEAD~\cite{kaisiyuan2020mead}) and in-the-wild sources (CelebV-HQ~\cite{zhu2022celebvhq}, HDTF~\cite{zhang2021flow}, and VoxCeleb2~\cite{chung18b_interspeech}), containing over 11.7k cleaned audio-video pairs with an average of over 15 seconds. We evaluate models on 200 video cases, each with a fixed duration of 10 seconds. More details are in \textbf{Appendix B.1}.

\mypara{Setup}  We use EMOCA with \texttt{EMOCA\_v2\_lr\_mse\_20} model. We use \texttt{wavlm-base-plus} and \texttt{emotion2vec\_plus\_base} to extract global audio embeddings and frame-wise emotion features from the input audio, respectively. Our integrated audio-rich backbone consists of 6 diffusion transformer blocks with a hidden size of 768, where each prediction head is implemented with 2 blocks. We train 3DXTalker for 100 epochs with a batch size of 128 and a frame length of 250 on two NVIDIA H100 GPUs. Optimization is performed with AdamW (learning rate $1\times10^{-4}$, weight decay 0.01) with a OneCycleLR scheduler. Flow-matching steps are set to 512 for training and 32 for inference. Finally, following prior works~\cite{peng2023emotalk, sun2024diffposetalk}, we apply a Savitzky–Golay filter to the predicted sequences, which further improves motion smoothness. More details are in \textbf{Appendix B.2}.

\mypara{Metrics}
We evaluate 3DXTalker using 9 metrics across four dimensions. Identity preservation is measured by MVE~\cite{fan2024unitalker} and CSIM~\cite{ghazouali2024csim}, where MVE reflects  3D geometric error and CSIM measures 2D identity consistency. Lip synchronization is assessed using LVE~\cite{faceformer2022}, LSEC and LSED~\cite{prajwal2020lip}, where LVE measures 3D vertex-level lip alignment, while LSEC and LSED quantify audio–2D visual sync quality through confidence and embedding-distance mismatch. Facial expression quality is evaluated using UFVE~\cite{stan2023facediffuser}, UFDD~\cite{xing2023codetalker}, and Emo-FID~\cite{kim2024deeptalk}; UFVE measures 3D upper-face geometry errors, UFDD examines 3D temporal smoothness of expression dynamics, and Emo-FID reflects 2D facial expression similarities. Head-pose motion is measured using beat alignment (BA) score~\cite{sun2024diffposetalk, siyao2022bailando}. Besides, we evaluate perceptual quality via a subjective User Study, reporting the mean rank (MR) based on participant preferences. Finally, FPS is recorded during inference for efficiency evaluation. We compute 3D metrics on meshes and 2D metrics on rendered videos.
Details are provided in \textbf{Appendix B.3}.

\begin{table*}[t]
\centering
\caption{Quantitative evaluation with baselines over 200 videos. ``ID Gen." denotes whether the model supports identity-referred generalization.  Our method achieves better performance on 3D-related metrics, demonstrating improved geometric accuracy and temporal coherence, while it performs well on most 2D perceptual metrics. The BA score assesses the rhythmic alignment of head motions with the driving audio. The MR is averaged over four rating dimensions in a study with 26 participants, each evaluating 20 test cases; lower is better. Throughput was measured on an NVIDIA RTX 5090 GPU.
The \textbf{best} results are highlighted in bold and the \underline{second-best} results are underlined.}
\label{tab:Main_Quantative}%
\setlength{\tabcolsep}{1.5pt}    
\renewcommand{\arraystretch}{1}
\scalebox{0.8}{
    \begin{tabular}{l c ccc ccc ccc c c c} 
        \toprule[1pt]
         &  &  \multicolumn{3}{c}{\textbf{Identity}} & \multicolumn{3}{c}{\textbf{Lip-Sync}}  & \multicolumn{3}{c}{\textbf{Expression}} & \multicolumn{1}{c}{\textbf{Pose}} & \multicolumn{1}{c}{\textbf{User Study}} & \multicolumn{1}{c}{\textbf{Efficiency}} \\
        \cmidrule(lr){3-5} \cmidrule(lr){6-8} \cmidrule(lr){9-11} \cmidrule(lr){12-12} \cmidrule(lr){13-13} \cmidrule(lr){14-14}
        \textbf{Model} & Predition
        & ID & MVE  & CSIM
        & LVE  & LSEC & LSED
        & UFVE & UFDD  & Emo-Fid 
        & BA  
        & MR
        & Throughput \\
        
        & Type
        & Gen. & ($\times 10^{-3}$) $\downarrow$ & $\uparrow$
        & ($\times 10^{-4}$) $\downarrow$ & $\uparrow$ & $\downarrow$
        & ($\times 10^{-3}$) $\downarrow$ & ($\times 10^{-4}$) $\downarrow$ & $\downarrow$
        & $\uparrow$
        & $\downarrow$
         & (FPS) $\uparrow$ \\
        \midrule[1pt]
        
        FaceFormer         &\textit{Vert}  &\ding{55} &4.704 &0.919 &1.801  &0.751 & 13.584 & 4.094 & 4.970 & 0.351 & 0 & 4.46 & \underline{74.640} \\
        CodeTalker     &\textit{Vert}  &\ding{55} &4.795 &0.920 &1.886  &0.777 & 13.311 & 4.123 & 3.783 & 0.374 & 0 & 4.63 & 64.997 \\
        SelfTalk        &\textit{Vert}  &\ding{55} &4.803 &0.921 &1.863  &0.843 & 13.385 & 4.106 & 3.659 & \underline{0.336}  & 0 & 4.45 & 9.083 \\
        FaceDiffuser &\textit{Vert}  &\ding{55} &4.752 &\underline{0.933} &1.624 &0.636 & \textbf{12.732} &4.173 & 3.567 &0.398 & 0 & 4.58 & 16.142 \\
        EMOTE     &\textit{Para}  &\ding{55} &2.883 &0.896 &1.539  &\underline{2.706} & 13.488 & 2.615 & \underline{2.974} & 0.359 & 0 & 4.45 & \textbf{91.417} \\
        DEEPTalk         &\textit{Para}  &\ding{55} &\underline{2.672} &0.915 &\underline{1.535}  &2.174 & 13.564 &\underline{2.306} & 3.146 & 0.355 & 0 & \underline{4.28} & 59.604 \\
        DiffPoseTalk  &\textit{Para}  &\ding{55} &2.877 &0.876 &1.698 &\textbf{4.021} & \underline{13.069} &2.453  &3.187 & 0.377 & \underline{0.338} & 4.94 & 18.795 \\
        Ours                                     &\textit{Para} &\ding{51} &\textbf{1.371} &\textbf{0.967} &\textbf{0.877} &1.674 & 13.330 & \textbf{1.113} & \textbf{1.317} & \textbf{0.259} & \textbf{0.404} & \textbf{4.22} & 69.497 \\
        
        \bottomrule[1pt]
        \end{tabular}
    }
\end{table*}

\subsection{Main Results}
\mypara{Qualitative Evaluation}
Figure~\ref{fig:QualitativeComparison} provides qualitative comparisons against several representative baselines (Full comparisons are presented in \textbf{Appendix C}). As shown in Figure~\ref{fig:QualitativeComparison} (a), 3DXTalker effectively preserves identity while observing shapes and details, producing meshes that closely match the reference image. In Figure~\ref{fig:QualitativeComparison} (b), our method achieves clearer mouth aperture alignment across diverse word syllables, demonstrating a tighter correspondence between mouth movements and speech. Figure~\ref{fig:QualitativeComparison} (c) highlights the model’s ability to generate finer emotional expressions, capturing subtle variations in facial dynamics as the speech evolves. Finally, Figure~\ref{fig:QualitativeComparison} (d) illustrates natural head-pose behaviors along with different camera movements generated by our 3DXTalker. 
Overall, these results show that 3DXTalker can synthesize diverse, dynamic, and expressive 3D talking videos.

\begin{table}[t]
    \centering
    \caption{Ablation results of 3DXTalker. ``AbsLatent” uses absolute latent instead of differential ones in Eq. (\ref{eq:abs}). ``w/o $\mathbf{A}_{\text{emo}}$" and ``w/o $\mathbf{A}_{amp}$" remove emotional and amplitude embeddings. The best results in \textbf{bold} and second-best \underline{underlined}.}
    \label{tab:ablation_qualitative}
    \setlength{\tabcolsep}{2.5pt}
    \renewcommand{\arraystretch}{1.3}
    \scalebox{0.85}{
        \begin{tabular}{lcccccccc}
        \toprule[1pt]
        \textbf{Model} 
        & LVE & MVE  & UFDD & UFVE & LSEC & LSED & CSIM \\
        & ($\times 10^{-4}$)$\downarrow$  & ($\times 10^{-3}$)$\downarrow$  & ($\times 10^{-4}$)$\downarrow$ & ($\times 10^{-3}$)$\downarrow$ &$\uparrow$ &$\downarrow$ &$\uparrow$  \\
        \midrule
        $\text{AbsLatent}$                & 1.242 & 1.558 & 1.479 & 1.212 & \textbf{1.794} & \underline{13.348} & 0.959 \\
        
        w/o $\mathbf{A}_{amp}$             & 0.971 & 1.417 & \underline{1.328} & \underline{1.133} & 0.240 & 14.116 & \textbf{0.975} \\

        w/o $\mathbf{A}_{emo}$             & \underline{0.926} & \underline{1.414} & 1.373 & 1.142 & 1.308 & 13.425 & 0.966 \\
        
        \textbf{Ours}                     & \textbf{0.877} & \textbf{1.371} & \textbf{1.317} & \textbf{1.113} & \underline{1.674} & \textbf{13.330} & \underline{0.967} \\
        \bottomrule[1pt]
        \end{tabular}
    }
\end{table}

\begin{figure}[t]
    \centering
    \includegraphics[width=0.8\linewidth]{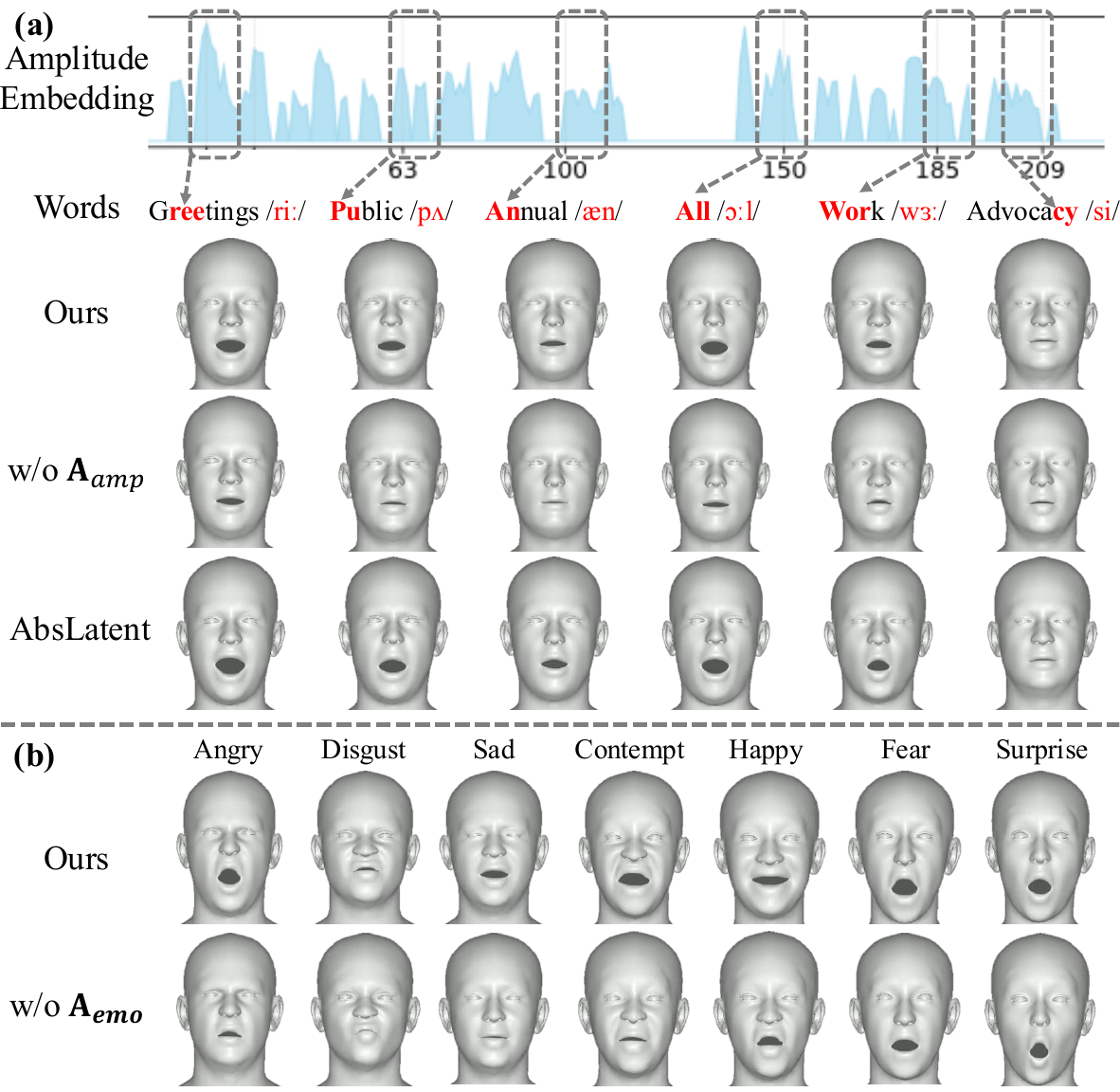}
    \caption{Visualizations of ablation results from Table~\ref{tab:ablation_qualitative}. (a) is conducted on the same audio. (b) extracts each emotion from corresponding videos at the same frame. More cases can be found in \textbf{Appendix F}.}

    \label{fig:ablation_visual}
\end{figure}

\mypara{Quantitative Evaluation}
We quantitatively compare our method with seven competitive baselines across identity, lip sync, emotional expression, head pose, and efficiency, as listed in Table \ref{tab:Main_Quantative}.
Our method performs well across the 3D-related metrics (MVE, LVE, UFVE, UFDD), suggesting accurate geometry reconstruction and stable temporal motion compared to the evaluated baselines. For 2D perceptual metrics, 3DXTalker attains competitive results on CSIM and Emo-FID, indicating consistent identity preservation and credible affective quality. The Lip-sync performance is strong overall, while LSEC and LSED remain challenging. We note that most 3D-based methods exhibit similarly limited improvements on these phoneme-sensitive metrics, except for FaceDiffuser. The BA score demonstrates that 3DXTalker can generate head-pose motions that are rhythmically synchronized with the input audio. In the user study, 3DXTalker obtains the lowest mean rank of 4.22, compared with 4.28 for DEEPTalk. This small difference indicates a lower average rank in the sampled evaluations, without establishing a statistically significant perceptual advantage. Finally, 3DXTalker achieves reasonable inference at 69.497 FPS.
These results collectively validate that 3DXTalker has strong potential for expressive generation, achieving a balanced trade-off between performance and efficiency.

\subsection{Ablation Study}
We conduct ablation experiments to evaluate the contribution of key components in 3DXTalker, with quantitative results listed in Table~\ref{tab:ablation_qualitative} and visual comparisons shown in Figure~\ref{fig:ablation_visual}.
Using absolute latent leads to the largest 3D errors and lower CSIM identity scores, indicating that our differential latent design in the dataset-curation pipeline helps separate identity from motion and is important for stable motion generation.
Removing $\mathbf{A}_{\text{amp}}$ increases CSIM since it reduces mouth-motion variation, which artificially enhances frame similarity (Figure~\ref{fig:ablation_visual} (a)). It also leads to higher 3D errors, confirming that amplitude cues provide essential constraints on realistic mouth aperture.
Likewise, removing $\mathbf{A}_{\text{emo}}$ degrades 3D performance across all geometric metrics and disrupts subtle emotional expression (Figure~\ref{fig:ablation_visual} (b)), underscoring the importance of frame-wise emotion features.
These findings validate our incorporation of frame-wise amplitude and emotion cues, forming audio-rich representations that strengthen speech-driven facial motions.

\begin{figure*}[t]
    \centering
    \begin{minipage}[t]{0.48\linewidth}
        \centering
        \includegraphics[width=\linewidth]{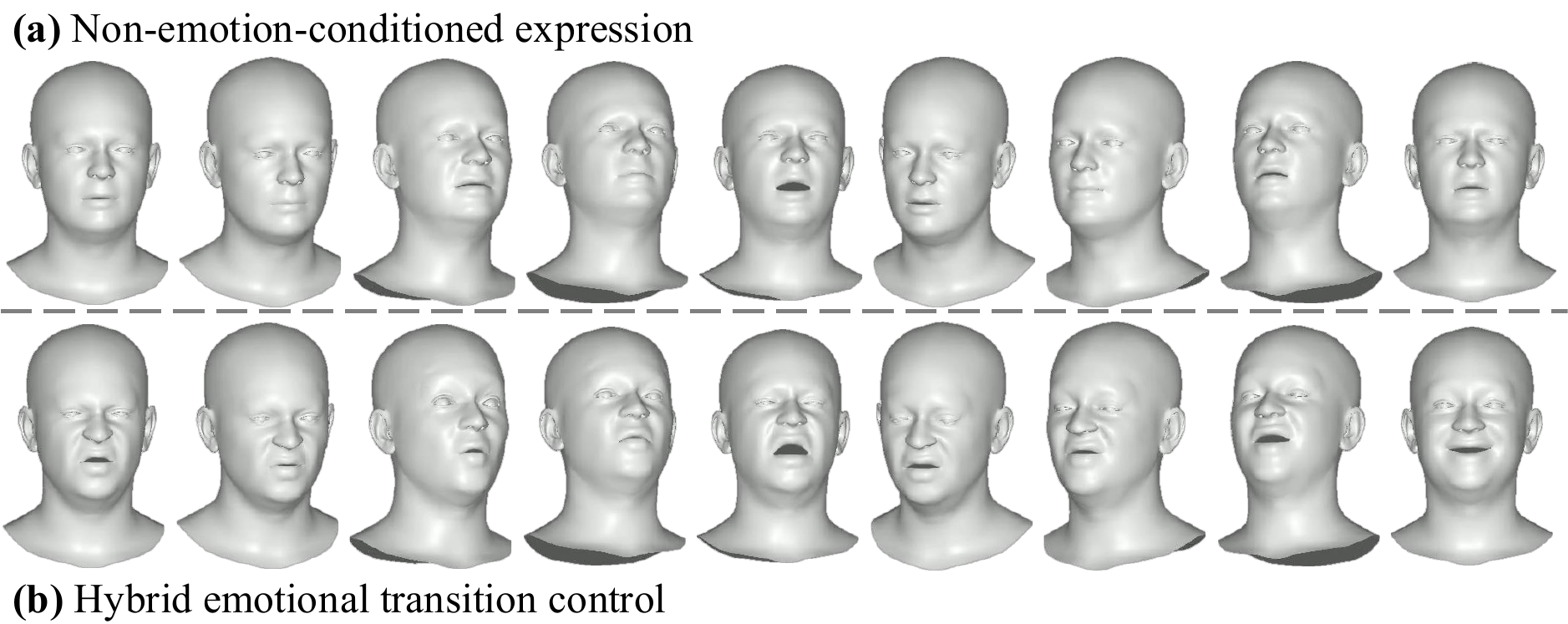}
        \caption{3DXTalker supports emotion control and seamless transitions between facial expressions. (a) shows the neutral talking-face state without guided emotion intervention, and (b) enables multiple emotion transitions among five emotion categories (angry $\rightarrow$ surprised $\rightarrow$ sad $\rightarrow$ contempt $\rightarrow$ happy).}
        \label{fig:emotion_control_analysis}
        
    \end{minipage}
    \hfill
    \begin{minipage}[t]{0.48\linewidth}
        \centering
        \includegraphics[width=\linewidth]{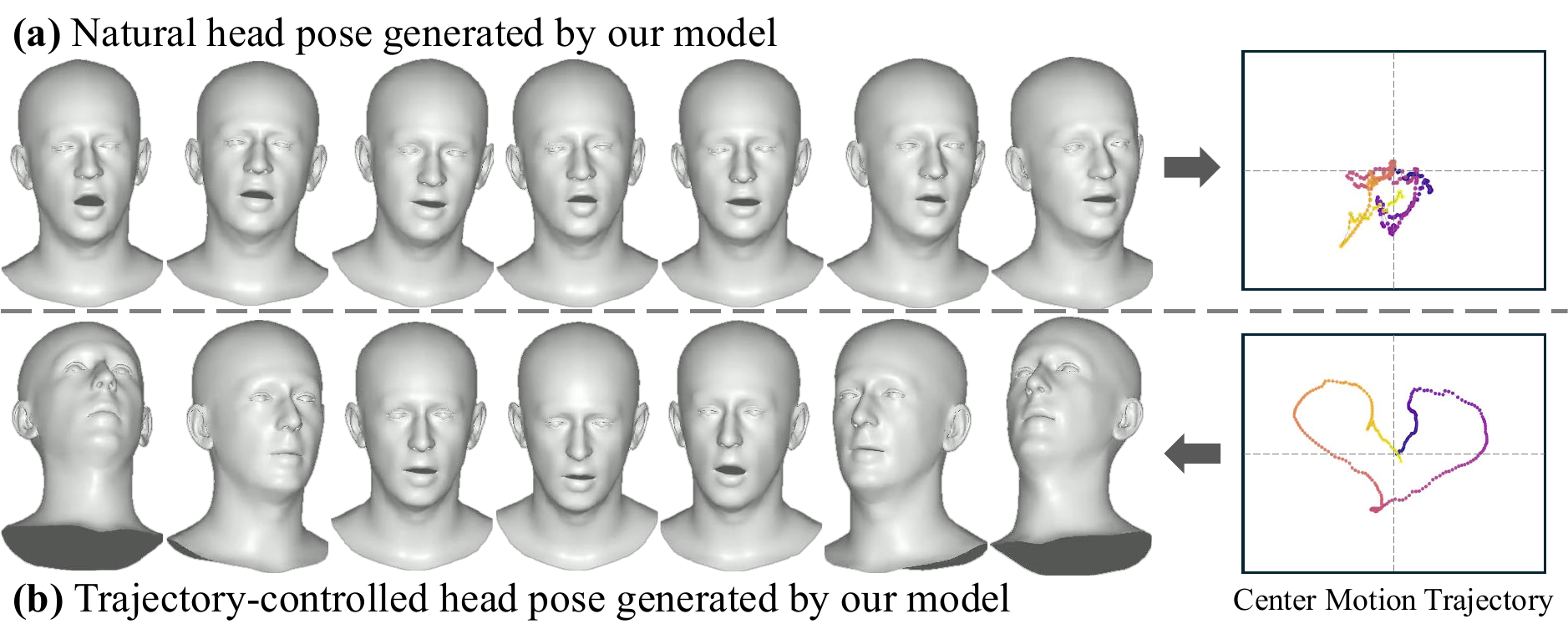}
        \caption{Our 3DXTalker supports two head-pose modes: (a) natural micro-movements learned from in-the-wild data, and (b) controllable head dynamics (with natural micro-movements) guided by a center motion trajectory. Trajectory colors indicate temporal progression (dark$\rightarrow$light). See \textbf{Appendix E.2} for more cases.}
        \label{fig:head_pose_analysis}
    \end{minipage}
\end{figure*}

\begin{table}[t]
    \centering
    \caption{Emotion-wise cosine similarity between generated and ground-truth FLAME expression parameters. ``$E_C$" denotes emotion control. Ang. = Angry, Con. = Contempt, Dis. = Disgust, Fea. = Fear, Hap. = Happy, Sad = Sad, Sur. = Surprise. Mesh visualizations and analyses are offered in \textbf{Appendix D}.}
    \vspace{5pt}
    \label{tab:emotion_similarity}
    \setlength{\tabcolsep}{6pt}
    \renewcommand{\arraystretch}{1.1}
    \scalebox{1}{
    \begin{tabular}{lccccccc}
    \toprule
    \textbf{Model} & Ang. & Con. & Dis. & Fea. & Hap. & Sad & Sur. \\
    \midrule
    DEEPTalk + $E_C$ & 0.637 & 0.738 & 0.602 & 0.686 & 0.721 & 0.659 & 0.640 \\
    Ours + $E_C$    & 0.947 & 0.935 & 0.953 & 0.933 & 0.921 & 0.939 & 0.938 \\
    \bottomrule
    \end{tabular}
    }
\end{table}

\begin{figure}[!htbp]
    \centering
    \includegraphics[width=1\linewidth]{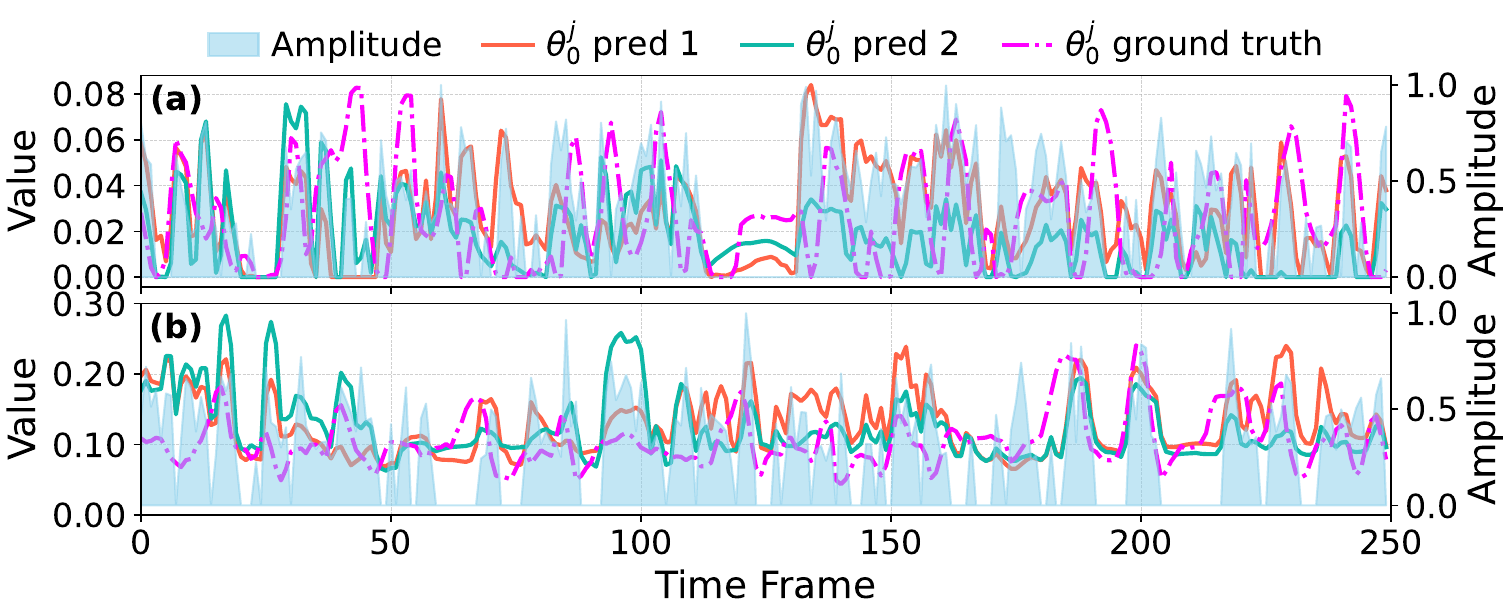}
    \caption{Curves for the ground truth and two predicted sequences, showing correlation with the amplitude-driven mouth aperture.}
    \label{fig:jaw_pose_analysis}
\end{figure}

\subsection{Analysis}
\mypara{Emotion Analysis} 
To further verify the model’s emotion controllability, we compute cosine similarity across seven emotions in the FLAME expression space, as listed in Table~\ref{tab:emotion_similarity}. 3DXTalker consistently achieves higher similarity scores than DEEPTalk, demonstrating superior emotion controllability and coherence. 
Figure~\ref{fig:emotion_control_analysis} shows temporal emotion controllability. In Figure~\ref{fig:emotion_control_analysis} (a) Without explicit emotion conditioning control, the talking head without emotion condition stays near a neutral state with limited affective variation. In contrast, Figure~\ref{fig:emotion_control_analysis} (b) shows that our hybrid transition control enables smooth, flicker-free transitions across multiple emotion categories while preserving identity-consistent geometry, indicating that the control signal modulates expression in a structured and continuous manner.

\mypara{Pose Analysis} 
We further analyze both jaw and head pose predictions. For jaw pose (Figure~\ref{fig:jaw_pose_analysis}), the predicted jaw-opening pose ($\boldsymbol{\theta}_0^{j}$) trajectories with two individual seeds exhibit temporal consistency with the ground truth and closely follow the audio-amplitude, demonstrating that amplitude conditioning effectively contributes to mouth aperture. Regarding head pose, as shown in Figure~\ref{fig:head_pose_analysis}, 3DXTalker maintains high stability and continuity across both supported modes. The results demonstrate that the model achieves both realistic pose generation and controllable dynamics without introducing abrupt jitter or pose discontinuities.

\FloatBarrier
\section{Conclusion}
In this paper, we present \textbf{3DXTalker}, an integrated flow-matching framework for expressive audio-driven 3D talking avatar generation. By combining scalable 2D-to-3D data integration, audio-rich representations, and plug-in semantic controllability, 3DXTalker jointly improves identity consistency, lip synchronization, emotional expression, and head-pose dynamics within a unified framework. This design enables expressive 3D talking avatars to better balance subject fidelity, speech alignment, emotional expressiveness, and  controllable spatial dynamics in a coordinated manner. Extensive experiments demonstrate that 3DXTalker achieves superior and more comprehensive performance in expressive 3D talking avatar generation, validating both its effectiveness and controllability. Overall, our work takes a practical step toward holistic expressivity in next-generation 3D avatar systems and provides a flexible foundation for future research on expressive digital humans.

\bibliographystyle{unsrt}  
\bibliography{main}

\newpage
\appendix
\begin{figure*}[t]
    \centering
    \includegraphics[width=0.9\linewidth]{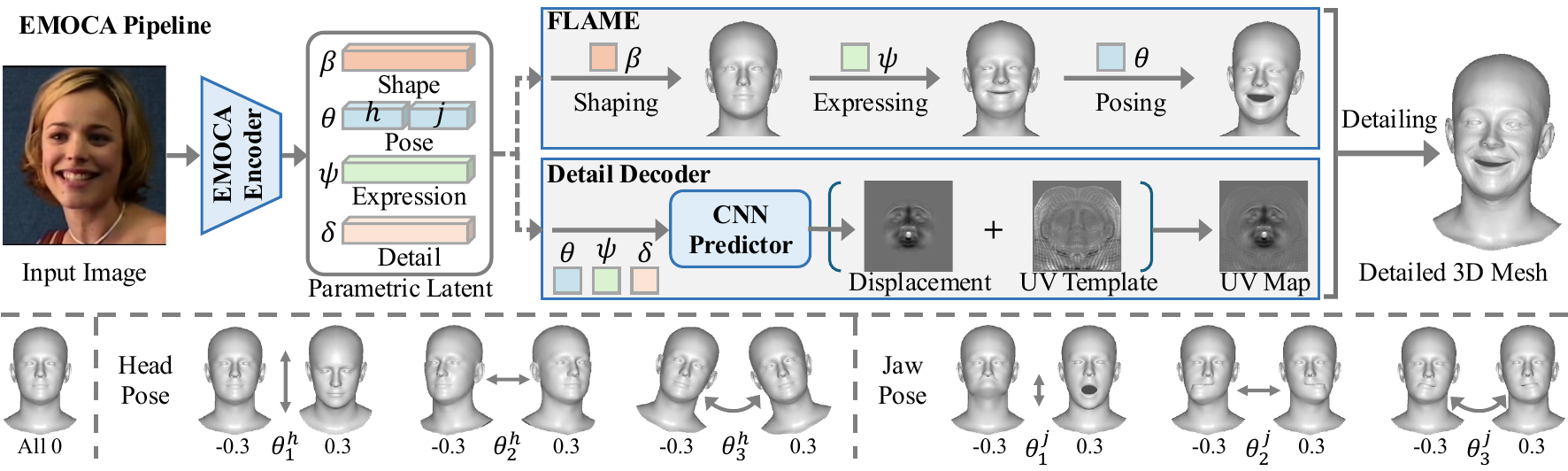}
    \caption{EMOCA modeling pipeline using the FLAME model.
    The encoder outputs parametric latent codes: $\boldsymbol{\beta}$ for facial shape, $\boldsymbol{\psi}$ for expression, $\boldsymbol{\theta}$ for head pose and jaw pose dynamics, and $\boldsymbol{\delta}$ for fine-grained appearance details (e.g., texture).
    We linearly vary $\boldsymbol{\theta}$ to demonstrate controllable changes in head and jaw pose, as illustrated in the bottom row.}
    \label{fig:PAE}
\end{figure*}

\section{EMOCA Preliminary}
\label{append:emoca}
To leverage the abundant identities, emotional styles, and motion patterns available in large-scale 2D video data, we adopt EMOCA~\cite{danvevcek2022emoca} as a parametric 2D-to-3D facial reconstruction model. As illustrated in Figure~\ref{fig:PAE}, EMOCA lifts each video frame into a structured FLAME parameter space, including shape, expression, pose, and detail codes, and further decodes them into a detailed 3D facial mesh. This formulation provides a controllable bridge from rich 2D visual observations to structured 3D facial representations, enabling scalable identity-aware modeling without requiring explicit 3D motion capture. It also offers a disentangled parameterization that is well suited for subsequent temporal motion modeling and controllable 3D avatar generation.

\begin{table*}[t]
\centering
\caption{Comparison of existing 3D talking-head datasets and our dataset. Our dataset provides the most comprehensive 3D fitting FLAME annotations, including identity, expression, pose, and detail, while also covering both lab and wild scenarios with speech and song audio. BS denotes blendshape coefficients, and ARKit refers to Apple’s facial motion capture parameterization.}
\label{tab:BenchmarkCompare}%
\setlength{\tabcolsep}{1pt}    
\renewcommand{\arraystretch}{1.2}
\scalebox{0.75}{
    \begin{tabular}{c|ccccccccc|cccc}
    \hline
    Dataset   & Representation & Dimension & Annotation & Source      & Audio        & Sequence & Duration & FPS & Subject & Identity & Expression & Pose & Detail \\ \hline
    Vocaset~\cite{VOCA2019}   & Vertices       & 5,023$\times$3   & 4D Scan    & Lab         & Speech       & 473      & 0.56h    & 60  & 12      & -        & -          & -    & -      \\
    BIWI~\cite{fanelli20103}      & Vertices       & 23,370$\times$3  & 4D Scan    & Lab         & Speech       & 238      & 0.33     & 25  & 6       & -        & -          & -    & -      \\
    Multiface~\cite{wuu2022multiface} & Vertices       & 6,172$\times$3   & 4D Scan    & Lab         & Speech       & 612      & 0.67h    & 30  & 13      & -        & -          & -    & -      \\
    MMFace4D~\cite{wu2023mmface4dlargescalemultimodal4d}  & Vertices       & 35,709$\times$3  & 4D Scan    & Lab         & Speech       & 35,904  & 36h      & 30  & 431     & -        & -          & -    & -      \\
    BEAT~\cite{liu2022beat}      & BS             & 52        & ARKit      & Lab         & Speech       & 2,508    & 76h      & 60  & 30      & -        & -          & -    & -      \\
    Song2face~\cite{iwase2020song2face} & BS             & 51        & ARKit      & Lab         & Song         &   -      & 1.93h    &  -   & 7       & -        & -          & -    & -      \\
    3D-ETF~\cite{peng2023emotalk}    & BS             & 52        & 3D fitting & Lab \& Wild   & Speech       & 3,479    & 6.97h    & 30  & 165     & -        & -          & -    & -      \\
    MMHead~\cite{wu2024mmhead}    & FLAME          & 56        & 3D fitting & Wild        & Speech       & 35,903   & 49h      & 25  & 2k+     & \ding{55}    & \ding{51}   & \ding{51}    & \ding{55}     \\
    TFHP~\cite{sun2024diffposetalk}      & FLAME          & 56        & 3D fitting & Wild        & Speech       & 1,052   & 26.5h    & 25  & 588     & \ding{55}    & \ding{51}   & \ding{51}    & \ding{55}  \\ \hline
    Ours      & FLAME          & 284       & 3D fitting & Lab \& Wild & Speech\&Song & 11,706   & 67.41h   & 25  & 2k+     & \ding{51}    & \ding{51}  & \ding{51}    & \ding{51}  \\ \hline
    \end{tabular}
    }
    \label{tab:dataset_comapare}
\end{table*}

\section{Implementation Details}
\subsection{Scalable Data Integration Pipeline}
\label{append:DataCuration}
To address the scarcity of training data and limited identity diversity in 3D talking avatar generation, we construct a large-scale corpus by integrating six diverse 2D video datasets. Specifically, we collect six widely used 2D talking video datasets, including three lab-controlled datasets (GRID~\cite{cooke2006audio}, RAVDESS~\cite{livingstone2018ryerson}, MEAD~\cite{kaisiyuan2020mead}) and three in-the-wild datasets (VoxCeleb2~\cite{chung18b_interspeech}, HDTF~\cite{zhang2021flow}, CelebV-HQ~\cite{zhu2022celebvhq}). The lab-controlled datasets provide high-quality recordings with articulated facial movements and emotional expressions, while the in-the-wild datasets introduce a broad range of identities, speaking styles, and natural head pose dynamics. A comprehensive comparison of our dataset with existing 3D talking-head datasets is presented in Table~\ref{tab:dataset_comapare}. Compared with prior vertex-based and blendshape-based datasets, our dataset is built in the FLAME parameter space, providing a more compact and semantically structured representation for controllable 3D facial animation. Moreover, among existing FLAME-based datasets, our dataset offers the most comprehensive annotations, jointly covering identity, expression, pose, and detail. It also maintains strong diversity in data sources, audio types, and subject coverage, making it a more suitable benchmark for expressive 3D talking avatar generation.

\begin{table*}[t]
    \centering
    \caption{Statistics of the constructed 3D talking-head dataset via our data integration pipeline.}
    \setlength{\tabcolsep}{2pt}
    \renewcommand{\arraystretch}{1.2}
    \scalebox{0.8}{
    \begin{tabular}{cccccccccc}
    \hline
    Dataset   & ID & Environment & Year & Raw Resolution  & Size & Subject & Total Duration (s) & Hours (h) & \multicolumn{1}{c}{Avg. Duration} (s/sample) \\ \hline
    GRID      & V0 & Lab         & 2006 & $720\times576$  & 6600              & 34      & 99257.81 & 27.57 & 15.04                         \\
    RAVDESS   & V1 & Lab         & 2018 & $1280\times1024$       & 613               & 24      & 10071.88 & 2.80  & 16.43                         \\
    MEAD      & V2 & Lab         & 2020 & $1920\times1080$       & 1969              & 60      & 42868.77 & 11.91 & 21.77                         \\
    VoxCeleb2 & V3 & Wild        & 2018 & 360P$\sim$720P  & 1326              & 1k+     & 21528.20 & 5.98  & 16.24                         \\
    HDTF      & V4 & Wild        & 2021 & 720P$\sim$1080P & 400               & 300+    & 55452.08 & 15.40 & 138.63                        \\
    Celebv-HQ & V5 & Wild        & 2022 & $512\times512$         & 798                & 700+     &13486.20 & 3.75  & 16.90                         \\ \hline
    \end{tabular}
    }
    \label{tab:Dataset_distribution}
\end{table*}

\begin{table}[h]
    \centering
    \caption{Dataset split for training and test.}
    \setlength{\tabcolsep}{5pt}
    \renewcommand{\arraystretch}{1.1}
    \scalebox{0.9}{
        \begin{tabular}{ccccc}
        \hline
        Dataset  &ID & Total Size & Training Set & Test Set \\ \hline
        GRID     &V0 & 6600       & 6570  & 30   \\
        RAVDESS  &V1 & 613        & 583   & 30   \\
        MEAD     &V2 & 1969       & 1939  & 30   \\
        VoxCeleb2 &V3 & 1326       & 1996  & 30   \\
        HDTF      &V4 & 400        & 350   & 50   \\
        Celebv-HQ &V5 & 798        & 768  & 30   \\
        \hline
        Summary   & & 11706      & 11506 & 200  \\ \hline
        \end{tabular}
    }
    \label{tab:Dataset_split}
\end{table}

To ensure consistency and quality across talking video sources, we apply a unified data preprocessing pipeline to filter outliers, following main steps below:
\begin{enumerate}[label=(\arabic*)]
\item \textbf{Duration Filtering.} Lab-controlled datasets contain high-quality recordings with rich expressions but are limited by short clip lengths (3--5 seconds). To facilitate temporal modeling, we concatenate clips sharing the same identity and emotion, yielding sequences of approximately 10--20 seconds. 
In contrast, as in-the-wild datasets typically feature longer durations, we simply filter out samples shorter than 10 seconds.

\item \textbf{Signal-to-Noise Ratio Filtering.} To remove clips compromised by strong background noise, music, or environmental interference, 
we compute the signal-to-noise ratio (SNR) for each audio segment, discarding samples with SNR below a predefined threshold. This step is critical for in-the-wild datasets, where recordings often contain crowd noise, reverberation, or microphone artifacts. 
The SNR filtering of speech signals ensures reliable cues for reliable amplitude extraction, emotion inference, and audio–visual synchronization. By suppressing acoustically corrupted samples at the preprocessing stage, we further improve the robustness and consistency of the downstream speech-driven facial motion generation pipeline.

\item \textbf{Language Filtering.}  We enforce linguistic consistency by filtering clips based on spoken language using Whisper~\cite{radford2022whisper}, discarding non-English samples or those with low detection confidence. We retain only English videos to reduce cross-lingual variation in phonetic structure and prosodic patterns, which would otherwise introduce additional difficulty for learning stable audio-visual alignment. This step helps provide more consistent supervision for lip synchronization, emotion modeling, and speech-driven facial dynamics. This choice is made to improve training consistency rather than to restrict the framework to English-only applications. Extending to multi-lingual datasets is expected to further improve generalization and enable broader applicability.

\item \textbf{Audio-Visual Sync Filtering.} To guarantee strict audio to visual alignment, we filter samples using SyncNet~\cite{Chung16a} to evaluate the temporal correlation between lip motion and the speech signal. We discard clips exhibiting high synchronization errors, as well as those containing abrupt scene cuts or off-screen speakers (e.g., voice-overs). Eliminating these misaligned pairs is essential, as they provide incorrect supervision signals that can significantly degrade lip-sync learning and hinder expression modeling.

\item \textbf{Resolution Normalization.}
To ensure visual resolution across diverse datasets, each video is first resized by matching its shorter side to 512 pixels while preserving aspect ratio, followed by a center crop to obtain frames at a unified resolution of $512 \times 512$. After cropping, all videos are re-encoded at 25 FPS with standardized RGB for consistent motion sampling.
This normalization step harmonizes data from sources with varying aspect ratios, camera qualities, and spatial resolutions. 
\end{enumerate}

After the above preprocessing steps, we obtain a high-quality talking video corpus with the data distribution listed in Table~\ref{tab:Dataset_distribution}. The final curated dataset spans six sources, covering both lab-controlled and in-the-wild environments, and provides diverse identities, emotional expressions, and head pose dynamics. Finally, we apply EMOCA  to encode all curated videos into the FLAME parameter space, which serves for the training and evaluation of the model.

We follow a standard protocol across all datasets, reserving a small portion from each source for evaluation. As shown in Table~\ref{tab:Dataset_split}, we use 11,506 videos for training and 200 videos for testing. The test set includes balanced samples from various data sources, ensuring that the evaluation addresses controlled, emotional, and in-the-wild scenarios. This split facilitates a thorough assessment of identity generalization, emotional expressivity, and head-pose dynamics under various real-world conditions.

\subsection{Setup Details}
\label{append:setup}

We use EMOCA V2 to lift 2D videos into FLAME representations, with the pretrained weight \texttt{EMOCA\_v2\_lr\_mse\_20}. For audio encoding, we adopt the pretrained models \texttt{wavlm-base-plus} and \texttt{emotion2vec\_plu\_base} to extract global audio embeddings and frame-wise emotion features from input audio, respectively. These modules provide the structured 3D facial parameters and audio-rich cues used by the generative backbone.

The integrated audio-rich backbone of 3DXTalker consists of 6 diffusion transformer blocks with a hidden size of 768. Each prediction head is implemented with 2 additional blocks to refine branch-specific outputs from the shared hidden representation. In this way, the shared backbone models global identity, audio, and temporal dependencies, while the prediction heads specialize in different output factors.

Experiments are conducted on two NVIDIA H100 GPUs. For training, 3DXTalker is optimized for 100 epochs using 250-frame sequences and a batch size of 128. We adopt AdamW with a learning rate of $1\times10^{-4}$ and a weight decay of 0.01, together with a OneCycleLR scheduler. The flow-matching module uses 512 steps during training and 32 steps during inference, which provides a favorable balance between trajectory fidelity and computational efficiency. Following prior works~\cite{peng2023emotalk,sun2024diffposetalk}, we further apply a Savitzky--Golay filter to the predicted sequences during post-processing to improve motion smoothness.

During training, the sequence length is set to 250 frames. At inference, longer sequences are generated via a sliding-window strategy. Because each window is defined relative to a shared reference frame, identity is implicitly recalibrated across windows. In our experiments, outputs remain stable without noticeable degradation for sequences up to 10 minutes under typical talking-head settings. Longer durations are likely feasible, but were not further explored due to computational cost, indicating a soft rather than strict practical limit on sequence length.

\begin{figure}[t]
    \centering
    \includegraphics[width=1\linewidth]{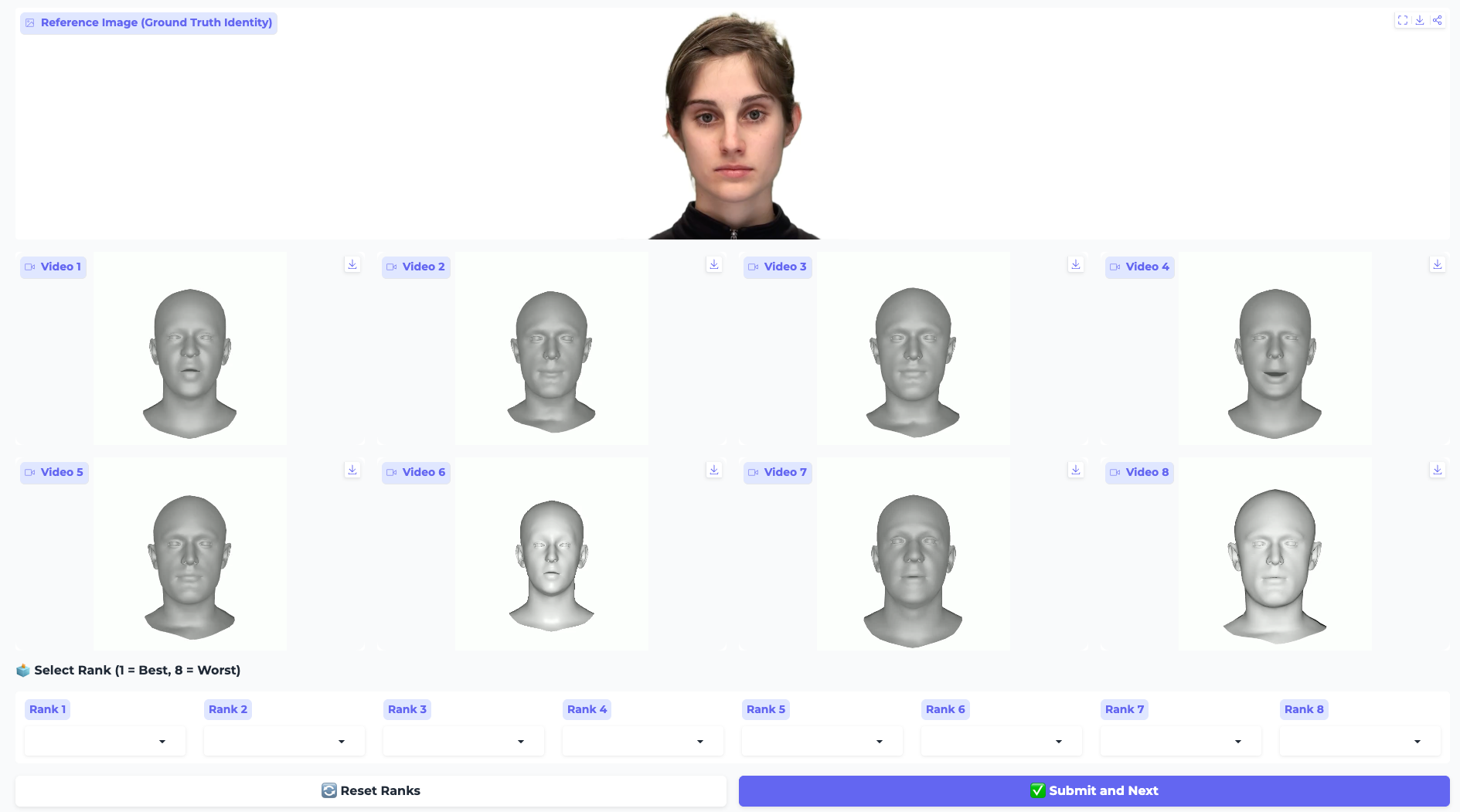}
    \caption{User study interface. Participants are presented with a ground-truth reference image (top) and eight anonymized video samples generated by all models. The videos are randomly shuffled to ensure a blind assessment.}
    \label{fig:user_study_ui}
\end{figure}

\subsection{Metrics Details}
\label{append:metrics}

We adopt 9 evaluation metrics across multiple levels, covering 3D geometry, 2D appearance,beat alignment score, and efficiency, to comprehensively assess 3D talking-avatar generation: 

\mypara{3D geometry}

\begin{itemize}[leftmargin=*, noitemsep, nolistsep]

\item Lip Vertex Error (LVE)~\cite{faceformer2022} measures lip synchronization by computing the mean Euclidean distance between the predicted and ground-truth lip-related mesh vertices; 
\item Upper Face Vertex Error (UFVE)~\cite{stan2023facediffuser} measures the mean Euclidean distance between predicted upper face mesh vertices and ground-truth; 
\item Upper Face Dynamics Deviation (UFDD)~\cite{xing2023codetalker} measures the deviation in facial dynamics for motion sequences between the predicted upper face mesh vertices and the ground-truth; Both UFVE and UFDD focus on the upper face area, specifically the forehead, eye region, and nose. 
\item Mean Vertex Error (MVE)~\cite{fan2024unitalker} evaluates geometric reconstruction accuracy by calculating the average Euclidean distance between corresponding vertices of the generated mesh and the ground truth across the entire head region.
\end{itemize}
\vspace{10pt}
\mypara{2D appearance}
\begin{itemize}[leftmargin=*, noitemsep, nolistsep]

\item 
Copula-based Similarity Metric (CSIM)~\cite{ghazouali2024csim} computes copula-based similarity by capturing the image features between the predicted video and ground truth per frame; 
\item Emotion Fréchet Distance (Emo-FID)~\cite{kim2024deeptalk} measures the similarity of emotional expressions between generated and ground-truth videos by computing the Fréchet distance between emotional embeddings extracted using the BEiT-Large model fine-tuned on AffectNet~\cite{tanneru2025beit_affectnet}; 
\item Lip-Sync Error Confidence (LSEC) and Lip-Sync Error Distance (LSED)~\cite{facetalkaudiodrivenmotiondiffusion} quantify audio–2D visual sync quality through confidence and embedding-distance mismatch, as followed by SyncNet~\cite{Chung16a}.
\end{itemize}

\mypara{Pose alignment}

\begin{itemize}[leftmargin=*, noitemsep, nolistsep]

\item Beat Alignment (BA) computes the average temporal distance between each audio beat
and its closest motion beat.
\end{itemize}

\mypara{User study}
\begin{itemize}[leftmargin=*, noitemsep, nolistsep]

\item We conduct a blind user study using the interface in Figure~\ref{fig:user_study_ui}. The study includes 26 participants, each evaluating 20 test cases randomly sampled from the 200-sequence test set. For each case, participants see a ground-truth reference image and eight anonymized videos generated by 3DXTalker and seven baselines, corresponding to 160 generated videos per participant. The display order is randomized independently for each case. Participants rank the eight videos from 1 (best) to 8 (worst) separately along four dimensions: identity consistency, lip synchronization, emotional expression, and head-pose dynamics. For each method, Mean Rank (MR) is the arithmetic mean of the ranks across all four dimensions, test cases, and participants; lower values indicate better overall perceptual quality.
\end{itemize}

\begin{figure*}[t]
    \centering
    \includegraphics[width=1\linewidth]{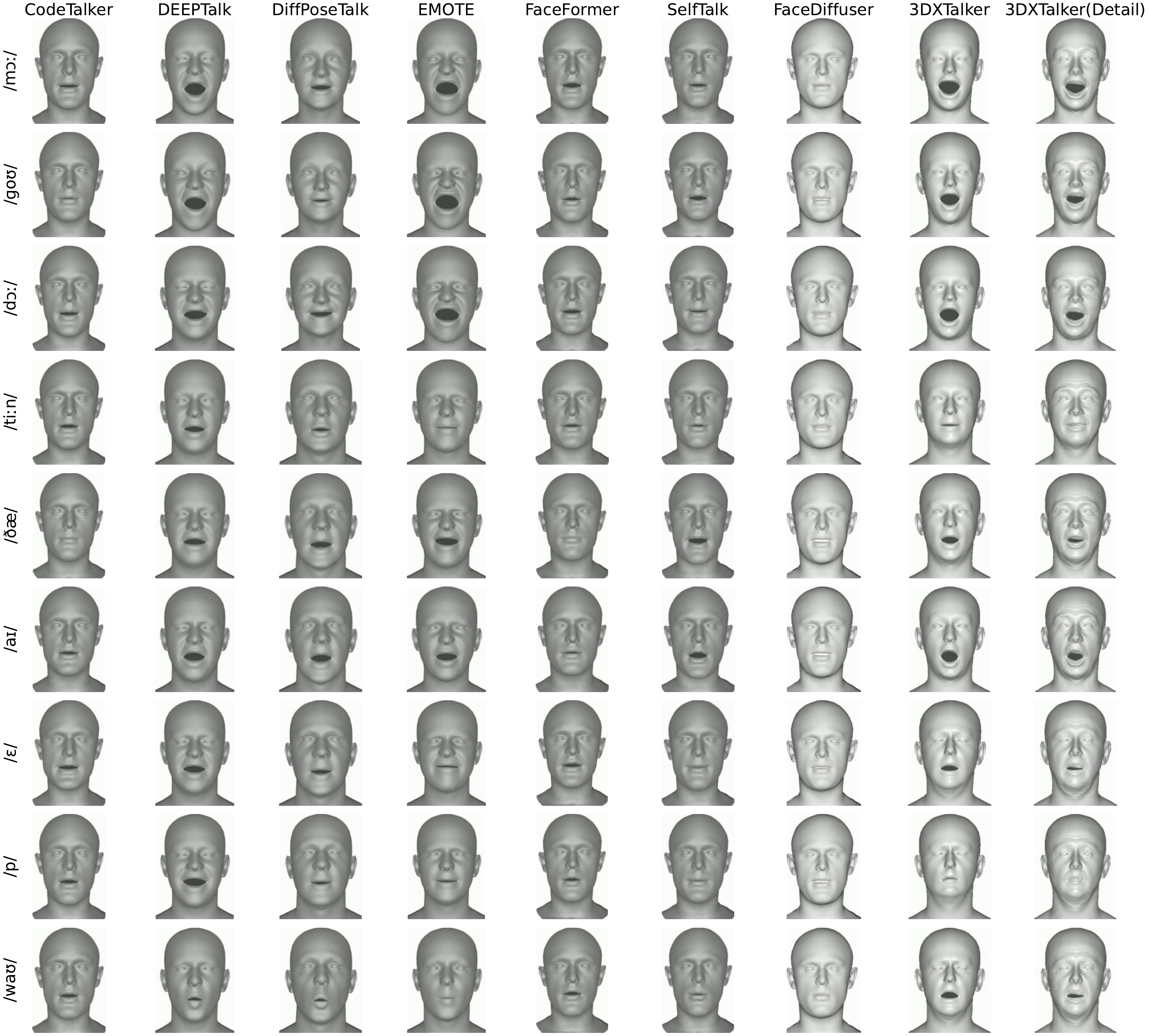}
    \caption{Visualization comparisons illustrating how mouth-aperture patterns align with phonetic symbols across models.}
    \label{fig:baseline_visualization}
\end{figure*}

\section{Full Baseline Comparisons}
\label{append:baseline}

We compare the performance of our 3DXTalker with seven representative baselines, as illustrated in Figure~\ref{fig:baseline_visualization}.

\begin{itemize}[leftmargin=*, noitemsep, nolistsep]

\item \textbf{FaceFormer~\cite{faceformer2022}}: employs an autoregressive transformer to predict 3D facial animation from speech, modeling temporal dependencies through sequential tokens.

\item \textbf{CodeTalker~\cite{xing2023codetalker}}: employs a discrete codebook representation to produce facial motions, enabling controllable audio–driven animation with compact latent tokens.

\item \textbf{SelfTalk~\cite{peng2023selftalk}}: employs a self-supervised commutative training scheme to learn 3D talking-face dynamics without paired data, enabling coherent audio–visual alignment through cycle-style consistency constraints.

\item \textbf{FaceDiffuser~\cite{stan2023facediffuser}}: applies diffusion modeling in a latent motion space to generate temporally coherent 3D facial animations conditioned on speech features.

\item \textbf{EMOTE~\cite{danvevcek2023emotional}}: disentangles speech content and emotion in a dual-branch architecture to drive 3D facial animation, enabling expressive emotion-aware motion generation from audio.

\item \textbf{DEEPTalk~\cite{kim2024deeptalk}}: predicts FLAME expression and jaw pose in a parametric subspace using audio-aligned diffusion, focusing on emotional expressiveness and articulation accuracy.

\item \textbf{DiffPoseTalk~\cite{sun2024diffposetalk}}: utilizes diffusion-based regression to estimate FLAME expression and 3D head pose from audio, generating articulated facial motions with controllable pose trajectories.
\end{itemize}

\begin{figure}[t]
    \centering
    \includegraphics[width=1\linewidth]{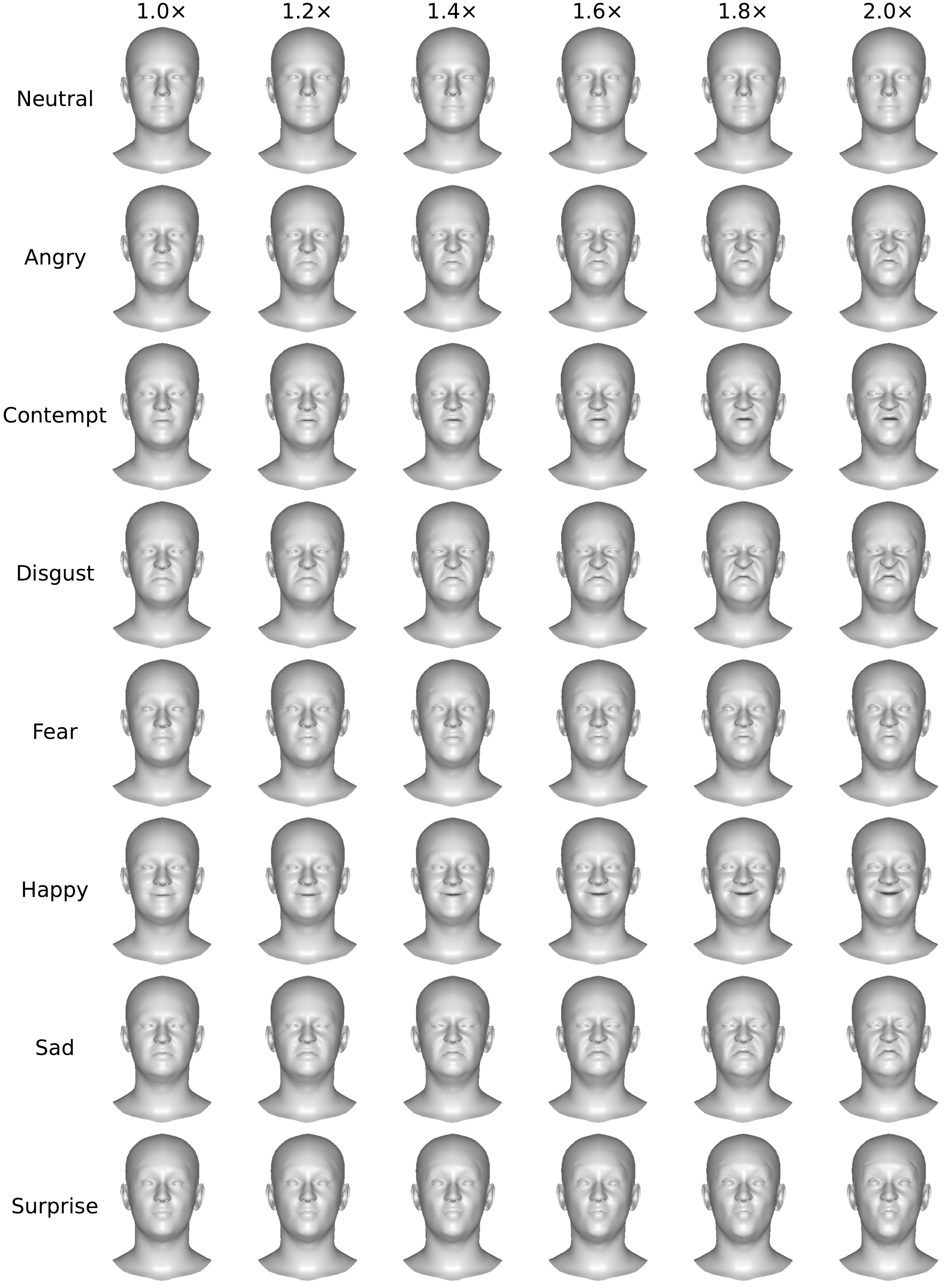}
    \caption{Neutral face (expression with zero vector) and seven emotion templates across six controllable intensity scales.}
    \label{fig:Emotion_Intensity}
\end{figure}

\begin{table}[h]
\centering
\caption{Procedure for extracting emotion templates from the MEAD dataset.}
\label{tab:emotion_template}
\renewcommand{\arraystretch}{1.25}
\setlength{\tabcolsep}{6pt}
\begin{tabular}{p{0.95\linewidth}}
\toprule
\textbf{Algorithm: Extraction of Emotion Expression Templates from MEAD} \\
\midrule
\textbf{Input:} MEAD expression-code sessions $\{\Psi_s\}$; emotion label $e$. \\
\textbf{Output:} Mean template $\bar{\psi}^{\,e}$. \\


\textbf{1.} Initialize frame set:  $\mathcal{X}_e \leftarrow \emptyset$. \\

\textbf{2.} \textbf{For each} session $s$ \textbf{containing emotion} $e$: \\
\quad \quad Load expression sequence $\Psi_s \in \mathbb{R}^{T_s \times 50}$. \\
\quad \quad Reshape to frame-level samples. \\
\quad \quad Update frame set:  
$\mathcal{X}_e \leftarrow \mathcal{X}_e \cup \{\Psi_s\}$. \\

\textbf{3.} Concatenate samples across all sessions:  
\[
    \mathcal{X}_e \in \mathbb{R}^{N_e \times 50}.
\]

\textbf{4.} Compute mean-based template:
\[
    \bar{\psi}^{\,e} = \frac{1}{N_e} \sum_{i=1}^{N_e} \mathcal{X}_e[i].
\]




\textbf{5.} \textbf{Return:} $\{\bar{\boldsymbol{\psi}}^{e}\}_{e=1}^7$. \\

\bottomrule
\end{tabular}
\end{table}

\section{Emotion Expression}
\label{append:emotion_expression}
\subsection{Semantic Emotion Control}
\label{append:emotion}
To enable explicit global emotion control in addition to audio-driven fine-grained dynamics, we derive emotion templates directly from the FLAME expression parameter subspace. These templates represent canonical expression directions for seven basic emotions and are used for emotion scaling and interpolation during inference. Furthermore, MEAD dataset provides videos recorded under controlled lighting and contains high-intensity emotion performances, making it suitable for learning expression prototypes.
Inspired by this, we obtain expression templates by analyzing the FLAME expression parameters ($\boldsymbol{\psi} \in \mathbb{R}^{50}$) extracted by EMOCA  from the MEAD dataset, with the full procedure detailed in Table~\ref{tab:emotion_template}. The extracted expression templates cover seven typical emotions (Angry, Contempt, Disgust, Fear, Happy, Sad, and Surprise). We further introduce a global scaling factor $\alpha\in\{1.0, 1.2,1.4,1.6,1.8,2.0\}$ to control emotion intensity, visualized in Figure~\ref{fig:Emotion_Intensity}.
During inference, we can adjust the global emotional tone by interpolating between the reference expression $\boldsymbol{\psi}_{ref}$ and the scaled emotion template:
\begin{equation}
\boldsymbol{\hat{\psi}}_{ref}^{\,e}
=
(1-\lambda)\,\boldsymbol{\psi}_{ref}
+
\lambda\,\alpha_e\, \bar{\boldsymbol{\psi}}^{e}.
\end{equation}
This yields seven categories of global emotion control, each with six adjustable intensities while preserving audio-driven local expression dynamics. 

\subsection{More Emotion Visualization Comparisons}
\label{app:more_emo_compare}

\begin{figure*}[t]
    \centering
    \includegraphics[width=1\linewidth]{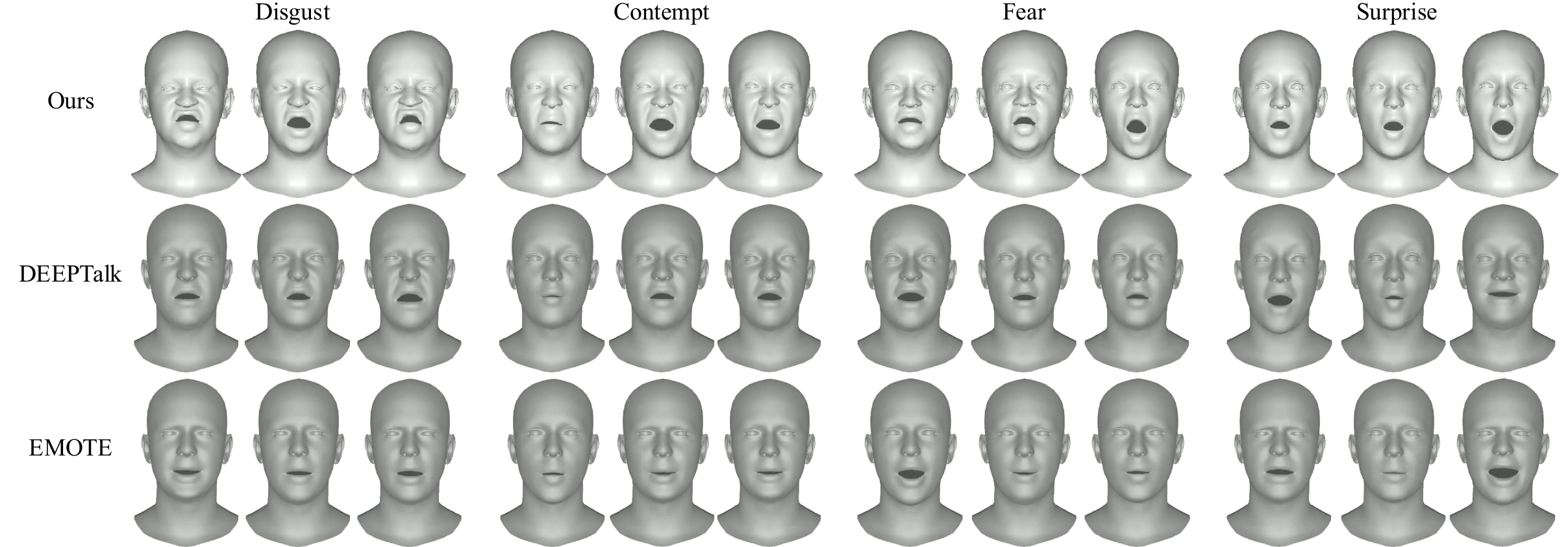}
    \caption{More qualitative comparisons of additional four emotion categories (Disgust, Contempt, Fear, Surprise) across representative baselines. 3DXTalker generates more expressive emotion patterns with clearer facial activations than DEEPTalk and EMOTE.}
    \label{fig:Emotion_cases}
\end{figure*}

To further demonstrate our model’s emotion expressivity, we further present qualitative comparisons across additional four representative emotion categories, as shown in Figure~\ref{fig:Emotion_cases}. Across all categories, our model produces more expressive and coherent facial deformations, demonstrating improved emotion disentanglement and modulation compared with DEEPTalk~\cite{kim2024deeptalk} and EMOTE~\cite{danvevcek2023emotional}.  In particular, for more intense emotions such as Disgust and Fear, our model produces stronger deformations around key facial regions, including the mouth, eyes, and brows. For Contempt and Surprise, it also maintains better coordination among local facial movements, resulting in more coherent overall expressions. In comparison, DEEPTalk and EMOTE often generate weaker or less distinctive emotional patterns. These observations further support that 3DXTalker provides stronger emotion disentanglement and more effective emotion modulation in expressive 3D talking avatar generation. These comparisons suggest that our framework better preserves category-specific emotional characteristics while maintaining facial coherence. This improvement is consistent with our design choice of combining disentangled FLAME-based modeling with frame-wise emotion-aware conditioning, which helps separate emotional modulation from identity and other motion factors.

\begin{figure}[t]
    \centering
    \includegraphics[width=1\linewidth]{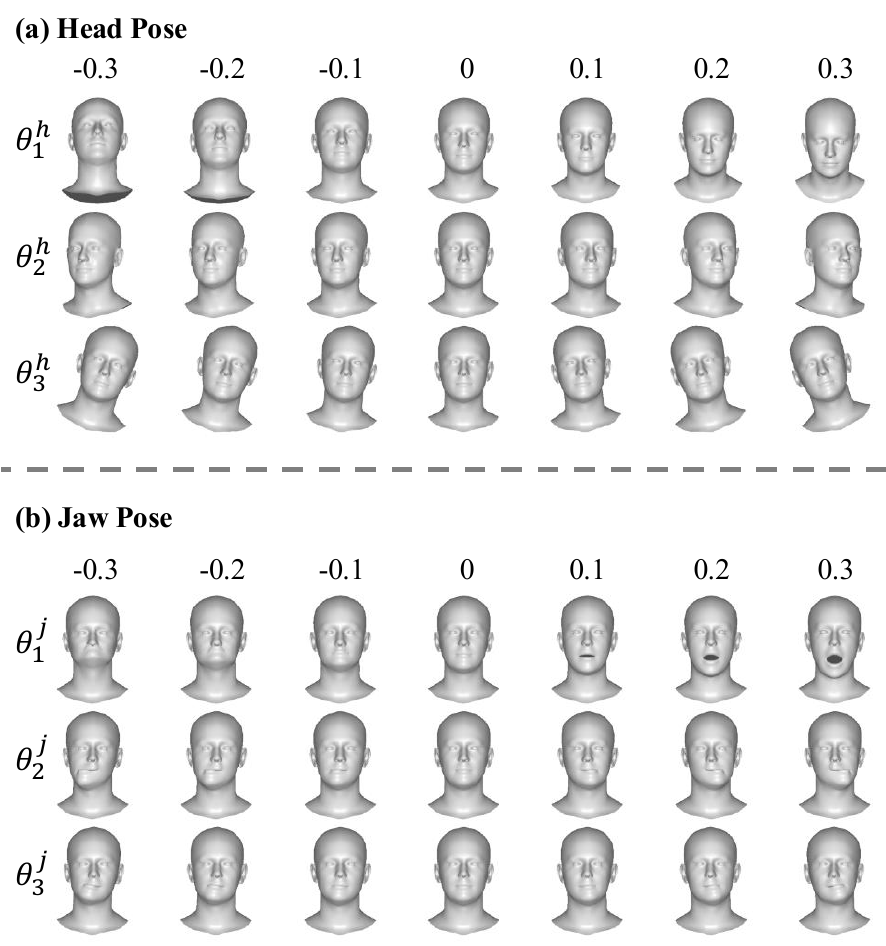}
    \caption{Head model visualization as the control parameter $\theta$ varies. (a) three parameters for head pose; (b) three parameters for jaw pose. They vary linearly with the control parameter.}
    \label{fig:theta}
\end{figure}

\section{Head Pose Dynamics}

\subsection{Semantic Pose Control}
\label{append:prompt}

To complement emotion semantic control, we further introduce semantic control over head-pose dynamics through an LLM-driven prompting strategy. Given the driving audio and a text prompt describing the desired presentation style, the language model generates a smooth and interpretable head-pose function for FLAME global pose parameters. The goal is to convert high-level semantic descriptions, such as ``energetic presentation'', ``subtle and calm'', or ``stage-presenting delivery'', into executable head-pose trajectories.
The prompt template used for this process is provided below.
By combining acoustic cues from the input waveform with the semantic intent specified in the prompt, the generated trajectory modulates head orientation over time in a smooth and controllable manner. Rather than replacing the model-learned natural head motion, the generated trajectory is superimposed onto it, thereby preserving realism while enabling semantically meaningful stylistic variation. As a result, 3DXTalker supports semantically meaningful head-pose variation while maintaining identity consistency and expression dynamics.

\begin{tcolorbox}
\ttfamily\small
\textbf{Instruction}: You are designing a global 
head-pose trajectory for a 3D talking avatar. The output must correspond to the FLAME global pose parameters ($\theta^g \in \mathbb{R}^{T\times3}$), represented as Rodrigues axis-angle vectors.
\textbf{INPUTS:} \\
- \textbf{audio}: a raw 1D NumPy array of the speech waveform (16 kHz). \\
- \textbf{style}: a text description of the desired singing or presentation style. (e.g., ``rhythmic and energetic”, ``calm and steady”, ``stage-presentation style”, ``passionate speech”).
\medskip
\textbf{Your task:} Produce a Python function \texttt{head\_pose\_func(T, audio)} that returns a NumPy array of shape [T, 3], where T is total number of frames, and each row is a Rodrigues axis angle rotation vector [pitch(y-axis), yaw(z-axis), roll(x-axis)] for the corresponding frame.
\medskip
\textbf{Requirements:} \\
- Directly analyze the audio waveform to detect: \\
\hspace*{1em}• energy peaks (emphasis). \\
\hspace*{1em}• low-energy regions (pauses). \\
\hspace*{1em}• rhythm / periodicity cues. \\[2pt]
- Use audio cues and a style prompt to control amplitude, pacing, and expressiveness of head motion. \\
- Produce smooth, continuous, interpretable trajectories. \\
- Ensure the resulting motion stays within natural 
FLAME global-pose ranges.
\end{tcolorbox}

\subsection{More Head-Pose Visualization Comparisons}
\label{append:pose}

\begin{figure}[t]
    \centering
    \includegraphics[width=0.95\linewidth]{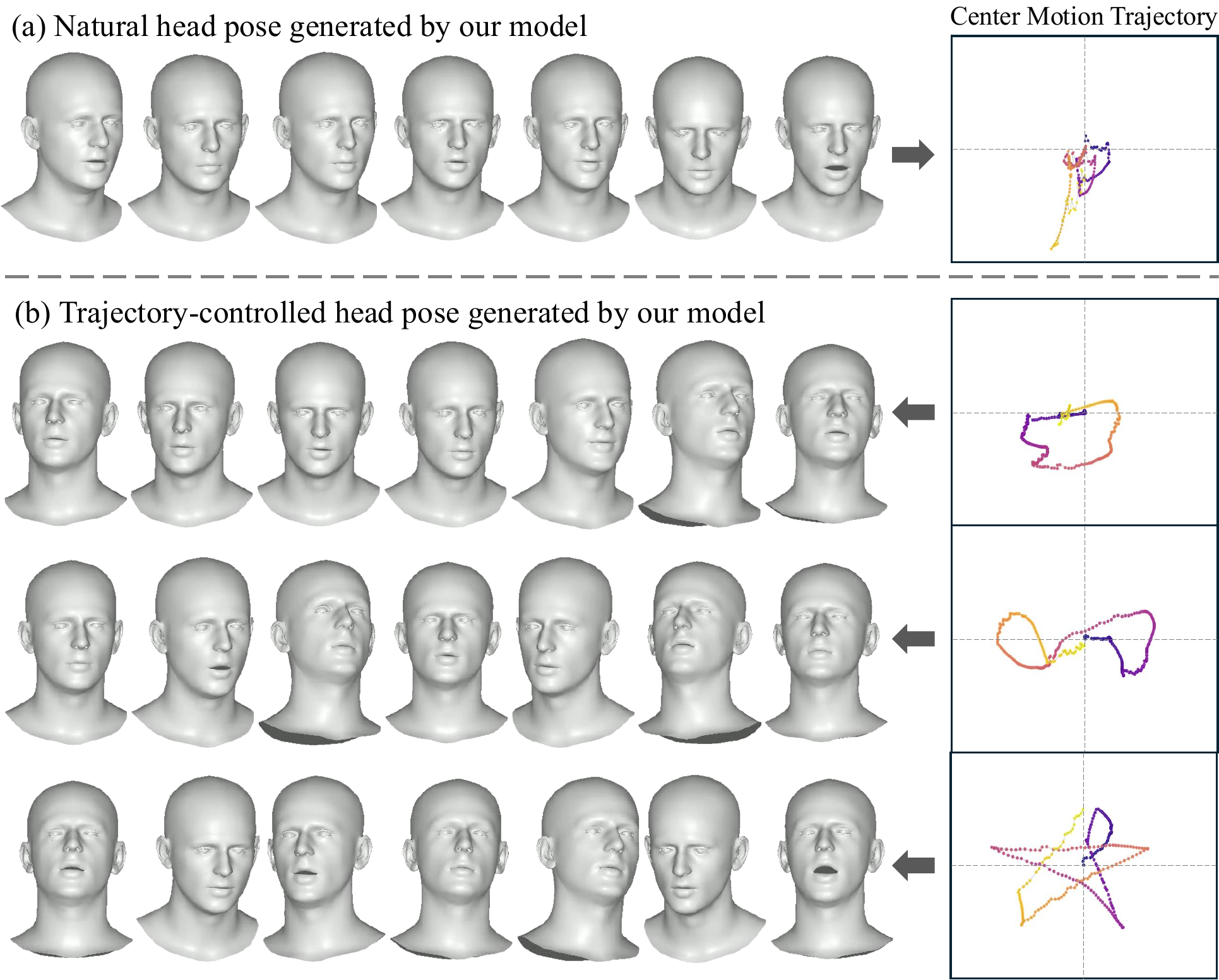}
    \caption{Comparison of head pose dynamics between the proposed natural micro-movement modeling and the trajectory-controlled head pose. By incorporating a center motion trajectory, the framework achieves both expressive flexibility and robust dynamic control. Trajectory colors indicate temporal progression (dark$\rightarrow$light).}
    \label{fig:head_pose_compare}
    \vspace{5pt}
\end{figure}

\begin{figure}[t]
    \centering
    \includegraphics[width=0.8\linewidth]{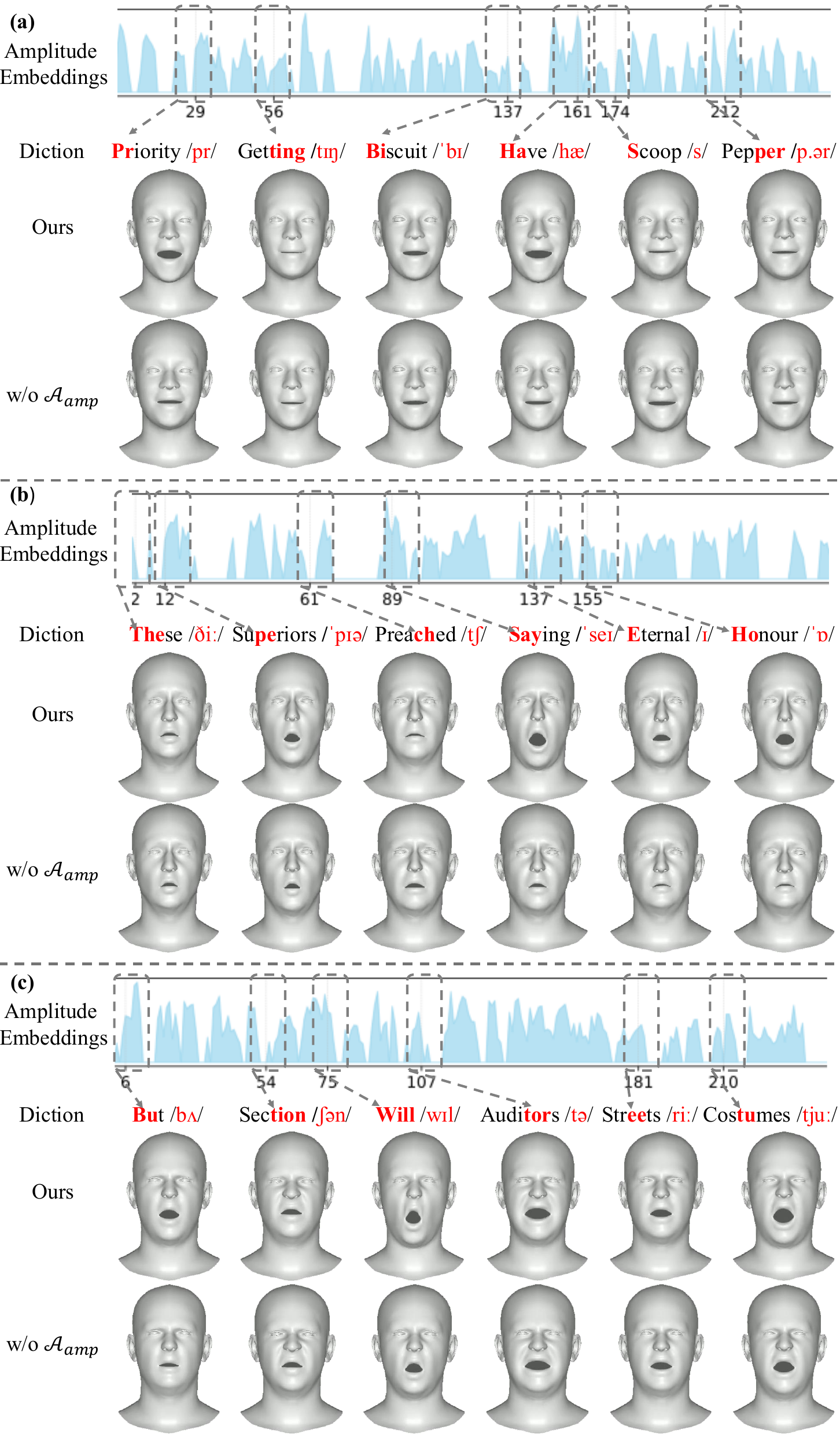}
    \caption{Amplitude analysis under different emotions, including (a) happy, (b) sad, and (c) angry.}
    \label{fig:amplitude_analysis}
\end{figure}

Expressive 3D talking avatars require coherent and flexible head pose motion in addition to accurate lip-sync and emotion-rich expressions. To better understand the controllability and behavior of the pose in the FLAME representation space, we linearly vary the $\boldsymbol{\theta}$ to visualize how individual dimensions influence head motion, as illustrated in Figure~\ref{fig:theta}. These variations confirm that each pose dimension corresponds to a meaningful and interpretable control component, such as pitch, yaw, roll, opening mouth, and pouting mouth, in a disentangled manner. Inspired by these disentangled pose behaviors, our 3DXTalker offers two complementary head-pose motion modes for diverse visual dynamics. First, the base model learns natural and realistic head sways directly from large-scale, in-the-wild datasets, producing subtle and coherent motions that align with the rhythm of speech. Second, 3DXTalker implements the center motion trajectory to control the head pose, along with the given direction and the deviation predicted by the model,  enables diverse motion patterns, such as energetic nods, stage-presentation style, or calm, minimal movements. Figure~\ref{fig:head_pose_compare} further shows the comparison of 3DXTalker with these two modes.
The comparison demonstrates that the default mode emphasizes natural and subtle realism, while the trajectory-controlled mode provides significantly more diverse and controllable motion styles.

\section{More Amplitude Analysis Cases}
\label{append:amplitude}

To further demonstrate the contribution of frame-wise amplitude embeddings, we provide more visualizations across three emotional conditions (happy, sad, and angry) as shown in Figure~\ref{fig:amplitude_analysis}. For each example, we align the audio amplitude envelope with the corresponding phonetic segments and compare the mouth aperture generated by our full model against the variant without amplitude embeddings (w/o $\mathbf{A}_{\text{amp}}$).
Across all emotions, our model produces mouth apertures that accurately reflect the local amplitude variations, resulting in natural changes in articulation strength. High-amplitude regions (e.g., stressed vowels and plosive consonants) correspond to visibly larger mouth apertures, while low-amplitude regions lead to more subtle movements. This demonstrates that the amplitude embedding effectively injects information about speech energy, enabling fine-grained control over lip dynamics.
In contrast, the w/o $\mathbf{A}_{\text{amp}}$ variant shows flattened or inconsistent articulation, where mouth openings vary weakly across phonemes and fail to reflect emphasis or prosodic changes. This effect is consistent across happy, sad, and angry expressions, indicating that amplitude cues are essential regardless of emotional state.
Overall, these analyzes confirm that amplitude embeddings enhance speech–mouth aperture alignment in 3D talking-head generation.

\begin{figure*}[t]
    \centering
    \noindent\makebox[\linewidth][r]{%
      \includegraphics[width=\linewidth]{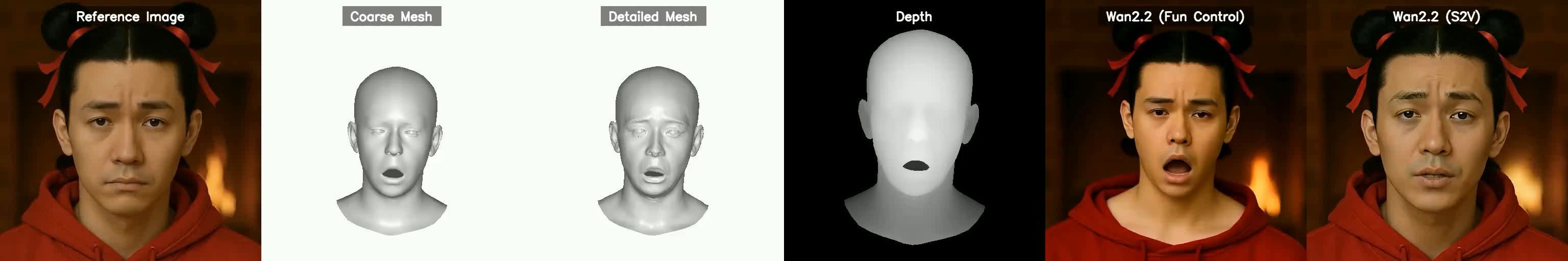}%
    }
    \noindent\makebox[\linewidth][r]{%
      \includegraphics[width=\linewidth]{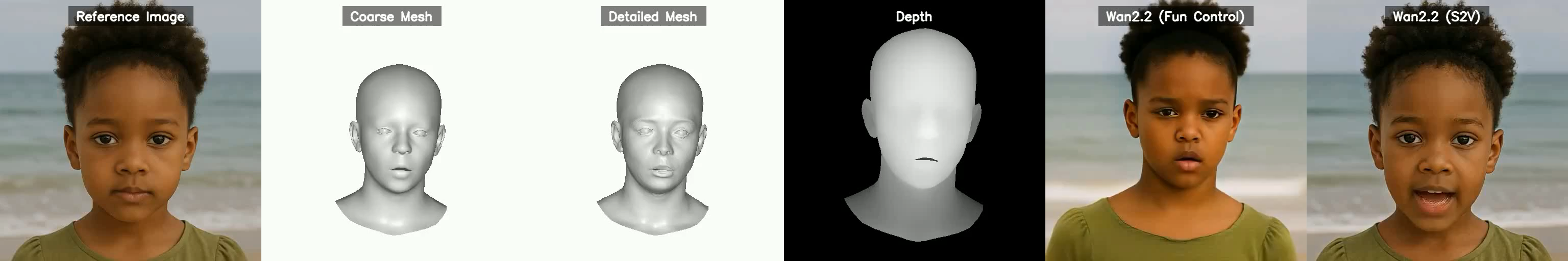}%
    }
    \caption{Wan2.2 rendering results comparison. Depth video are extracted from 3D mesh sequences generated by our 3DXTalker. Wan2.2 (Fun Control) synthesizes talking videos by conditioning on both the depth video and the reference image, while Wan2.2 (S2V) generates results directly from audio and the reference image. Both Fun Control and S2V models adopt text prompts to guide the generation process.}
    \label{fig:wan_render}
\end{figure*}

\section{Downstream Application: Wan 2.2}
\label{app:WanRender}
We employ ComfyUI to achieve texture mapping on the talking avatars generated by our model, using the Wan 2.2~\cite{wan2025}. Wan 2.2 is a versatile video generation framework that supports text-to-video, image-to-video, and video-to-video synthesis. To align with 3DXTalker's needs, we adopt two variants, Wan2.2-Fun-Control and Wan2.2-Speech-to-Video (S2V).
Fun-Control enables fine-grained video control through depth, pose, and edge guidance, complemented by LLM-generated prompts. Specifically, we render a depth video from our 3D mesh sequence and feed it into Fun-Control to drive head pose, lip movement, and emotion-consistent facial dynamics. Meanwhile, the reference image provides identity cues, ensuring that the synthesized video maintains the subject’s visual characteristics, while the input audio provides synchronized speech.
In contrast, S2V is an audio-driven video generation model that conditions on text prompts, reference images, and speech signals. It specializes in transforming a static portrait and audio into a synchronized talking video. However, unlike Fun-Control, S2V does not utilize depth guidance, which makes it less reliable in producing accurate head-pose dynamics and temporally coherent motion.

Figure~\ref{fig:wan_render} presents the rendered results produced using the Wan 2.2 model. Wan 2.2 requires two types of prompts: a positive prompt, which directs the model toward desired visual and motion characteristics, and a negative prompt, which constrains unwanted artifacts or behaviors during generation. In our setup, we vary only the positive prompts to tailor each video to its intended scene or stylistic effect while keeping the default negative prompt unchanged. For the full-control setting, the positive prompts are used together with the rendered depth videos and reference images to encourage identity-consistent and visually coherent talking video synthesis. The positive prompts for Figure~\ref{fig:wan_render} are presented below:
\begin{itemize}
    \item \ttfamily\small A realistic young Asian man is singing. His hair is styled in two rounded buns with red ribbons. He wears a red hoodie. He looks directly at the camera with a subtle emotional expression. Warm, fire-lit background.
    \item \ttfamily\small A realistic young African girl angrily speaking, with a tense brow, narrowed eyes, and tightly pressed lips. She has short natural curly hair and wears a green shirt. Strong front-facing expression, intense eye contact. Beach background with ocean and sky. Natural blinking and looking at the camera.
\end{itemize}

As shown in Figure~\ref{fig:wan_render}, both Fun Control and S2V effectively preserve the identity and background of the reference image. Fun Control transfers identity cues to a depth-conditioned mesh. However, the reliance on a constructed depth video imposes geometric constraints, resulting in an identity that is similar to the reference. Nevertheless, it provides stable identity across the sequence and accurately follows depth-guided head motion, lip movements, and emotional cues, resulting in natural dynamics, precise lip synchronization, and coherent temporal behavior. In contrast, S2V achieves closer identity fidelity due to its 2D appearance-based generation. However, its lack of geometric grounding results in poor audio–lip synchronization and minimal head-pose variation, producing rigid and temporally inconsistent outputs. Overall, the depth-conditioned Fun Control pipeline provides stronger geometric and temporal consistency, enabling high-fidelity head motion and lip articulation that purely 2D methods cannot achieve.

In the future, to further narrow the realism gap between 3D avatars and state-of-the-art 2D talking-head models, a promising direction is to incorporate a neural rendering module on top of the predicted FLAME geometry. By learning view-dependent appearance, fine-grained facial details, and realistic skin reflectance, such a neural renderer can transform mesh outputs into photorealistic renderings while preserving the controllability of our parametric model. This hybrid geometry–appearance approach has the potential to deliver high-fidelity visual quality comparable to 2D methods without sacrificing explicit 3D structure or editing flexibility.

\section{Future Work}

\mypara{Discussion on EMOCA Modeling}
As the upstream 2D-to-3D reconstruction module, EMOCA may introduce errors under challenging conditions such as occlusion, motion blur, extreme poses, or low-resolution facial regions. These inaccuracies can affect the recovered FLAME parameters, including shape, expression, pose, and detail, and may propagate to downstream motion generation. In particular, identity-related estimation errors may weaken identity consistency, while noisy expression or pose parameters may reduce motion stability.
Despite these limitations, EMOCA currently provides one of the most comprehensive monocular pipelines for jointly estimating identity, expression, pose, and detail parameters in a unified framework, making it well-suited for our task setting. While our pipeline incorporates standard data filtering and temporal smoothing steps that can partially reduce high-frequency artifacts, the reconstruction accuracy remains fundamentally constrained by the upstream lifting process.
Improving the robustness of 2D-to-3D reconstruction, especially under challenging visual conditions, represents an important direction for future work.

\mypara{Discussion on Dataset}
Although our training dataset is already relatively large-scale, the achievable level of fine-grained expressivity remains influenced by the diversity and quality of available data. Subtle speech-dependent facial dynamics, such as nuanced articulation patterns and culturally influenced expression styles, require sufficiently rich supervision in both audio and parameter spaces. Expanding the dataset with more fine-grained motion variations and broader linguistic diversity, particularly non-English speech data, could further improve the model’s ability to capture subtle expression differences and generalize across identities and speaking styles. Therefore, the upper bound of expressive controllability is determined not only by the representation capacity but also by the diversity and coverage of training data. Exploring richer multi-lingual and fine-grained motion datasets remains an important direction for future work to further enhance expressive fidelity.

\end{document}